\documentclass{article}

\usepackage[final]{neurips_2021}

\usepackage[utf8]{inputenc} % allow utf-8 input
\usepackage[T1]{fontenc}    % use 8-bit T1 fonts
\usepackage{url}            % simple URL typesetting
\usepackage{booktabs}       % professional-quality tables
\usepackage{amsfonts}       % blackboard math symbols
\usepackage{nicefrac}       % compact symbols for 1/2, etc.
\usepackage{microtype}      % microtypography
\usepackage{xcolor}         % colors

\usepackage{array}
\usepackage{graphicx}
\usepackage{clrscode}
\usepackage{enumitem}
\usepackage{multirow}
\usepackage{multicol}
\usepackage{float}
\usepackage{color}
\usepackage{amsopn}
\usepackage{mathrsfs}
\usepackage{mathtools}
\usepackage{amsmath}
\usepackage{arydshln}
\usepackage{blkarray}
\usepackage{courier}
\usepackage{rotating}
\usepackage{bm}
\usepackage{ragged2e}
\usepackage{setspace}
\usepackage{subfigure}
\usepackage{caption}
\usepackage{comment}
\usepackage{enumitem}
\usepackage{tabularx}
\usepackage{adjustbox}
\usepackage{hyperref}

\usepackage[linesnumbered,ruled,lined]{algorithm2e}
\usepackage{algorithmicx}
\usepackage{algpseudocode}
\usepackage{amsmath}

\definecolor{myblue}{rgb}{0, 0, 1}
\begin{document}
\title{Large Language Models Empowered Agent-based Modeling and Simulation: A Survey and Perspectives}

\author{%
	Chen Gao \quad Xiaochong Lan \quad Nian Li \quad Yuan Yuan \quad Jingtao Ding \quad Zhilun Zhou  \\ \textbf{Fengli Xu} \quad \textbf{Yong Li} \\
	Tsinghua University, Beijing, China \\
	\texttt{\{chgao96, fenglixu, liyong07\}@tsinghua.edu.cn}
}
\maketitle
\begin{abstract}
Agent-based modeling and simulation has evolved as a powerful tool for modeling complex systems, offering insights into emergent behaviors and interactions among diverse agents. Integrating large language models into agent-based modeling and simulation presents a promising avenue for enhancing simulation capabilities. 
This paper surveys the landscape of utilizing large language models in agent-based modeling and simulation, examining their challenges and promising future directions. 
In this survey, since this is an interdisciplinary field, we first introduce the background of agent-based modeling and simulation and large language model-empowered agents. We then discuss the motivation for applying large language models to agent-based simulation and systematically analyze the challenges in environment perception, human alignment, action generation, and evaluation.
Most importantly, we provide a comprehensive overview of the recent works of large language model-empowered agent-based modeling and simulation in multiple scenarios, which can be divided into four domains: cyber, physical, social, and hybrid, covering simulation of both real-world and virtual environments.
Finally, since this area is new and quickly evolving, we discuss the open problems and promising future directions.
\end{abstract}

\maketitle

\section{Introduction} 
Simulation, as a computational tool, encompasses the emulation of real-world processes or systems by employing mathematical formulas, algorithms, or computer-generated representations to imitate their behaviors or characteristics. Agent-based modeling and simulation focuses on modeling complex systems by simulating individual agents and their interactions within an environment~\cite{macal2005tutorial}. It operates by assigning specific behaviors, attributes, and decision-making capabilities to these agents, enabling the examination of emergent phenomena resulting from agents' interactions and environment dynamics. The significance of simulation spans various domains, serving as a valuable tool for understanding, analyzing, and predicting intricate phenomena that might be impractical or impossible to observe directly in real life. It facilitates experimentation, hypothesis testing, and scenario analysis, offering insights into systems' behaviors under diverse conditions and aiding in decision-making processes across fields like economics, biology, sociology, and ecology. 
The capacity to acquire and use language is a key aspect that distinguishes humans from other beings~\cite{hauser2002faculty}.
The advent of Large Language Models (LLMs) represents a recent milestone in machine learning, showcasing immense capabilities in natural language processing tasks and textual generation~\cite{zhao2023survey}. Leveraging their formidable abilities, LLMs have shown promise in enhancing agent-based simulations by enabling more nuanced and realistic representations of agents' decision-making processes, communication, and adaptation within simulated environments. Integrating LLMs into agent-based modeling and simulation holds the potential to enrich the fidelity and complexity of simulations, potentially yielding deeper insights into system behaviors and emergent phenomena for the following reasons:

First, the LLM agent can adaptively react and perform tasks based on the environment without predefined explicit instructions~\cite{autogpt, babyagi}. 
Second, the LLM agent has strong intelligence to respond like a human and even actively take actions with self-oriented planning and scheduling~\cite{wang2023survey,xi2023rise}. The action space of the LLM agent is neither limited to texts, for which the tool usage and internal action module allow the agent to take various actions~\cite{schick2023toolformer}.
Last, the LLM agent can interact and communicate with humans or other AI agents~\cite{park2023generative}.
With the above three strengths, LLM agents have been embraced for usage in a wide array of areas~\cite{park2022social,li2023you,li2023quantifying,park2023generative,kovavc2023socialai,lin2023agentsims,kovavc2023socialai,gao2023s,jinxin2023cgmi, boiko2023emergent,bran2023chemcrow}. From this perspective, it is clear that LLM agents can serve as a new paradigm for simulation, bestowing agents with human-level intelligence.

As a result of the massive potential of LLM agents, there has recently been a boom in research efforts in this area. However, as yet, there is no survey that systematically summarizes the relevant works, discusses the unresolved issues, and provides a glimpse into important research directions. In this survey, we analyze why large language models are essential in the fundamental problem of simulation, especially for agent-based simulation. After discussing how to design agents in this new paradigm, we carefully and extensively discuss and introduce the existing works in various areas, most of which have been published recently.
The contribution of this survey can be summarized as follows.
\begin{itemize}[leftmargin=*]
	\item We take the first step to review the existing works of large language model-based agent modeling and simulation.
		We systematically analyze why large language models can serve as an advanced solution for agent-based modeling and simulation compared with existing approaches. Specifically, we first extensively explain the requirements of the agent capability for agent-based modeling and simulation from four aspects: autonomy, social ability, reactivity, and pro-activeness. Then, we analyze how large language models address these challenges, including perception, reasoning and decision-making, adaptivity, and heterogeneity.
		\item We divide the agent-based modeling and simulation into four domains, physical, cyber, social, and hybrid, which can cover the mainstream simulation scenarios and tasks, after which we present the relevant works, providing a detailed discussion about how to design the simulation environment and how to build simulation agents driven by large language models.
	\item In addition to the existing works in this new area, we discuss four important research directions, including improving the simulation of scaling up, open simulation platform, robustness, ethical risks, etc., which we believe will inspire future research.
\end{itemize}

\section{Background}
In this section, we will first introduce the background of agent-based modeling and simulation, and large language models-empowered agents.

\subsection{Agent-based Simulation} 
\subsubsection{Basic concepts of agent-based simulation}
Agent-based simulation captures the intricate dynamics inherent in complex systems by concentrating on individual entities referred to as agents~\cite{macal2005tutorial}.
These agents are heterogeneous, with specific characteristics and states, and adaptively behave according to context and environment, 
, making decisions and taking actions~\cite{elsenbroich2014agent}.
The environment, whether static or evolving, introduces conditions, instigates competition, defines boundaries, and occasionally supplies resources influencing agent behaviors~\cite{cipi2011simulation}.
The interaction includes interactions with both the environment and other agents, and the goal is to mirror the behaviors in reality based on predefined or adaptive rules~\cite{elliott2002exploring,macal2005tutorial}. To summarize, the basic components of agent-based simulation include:

\textbf{Agents.} Agents are the fundamental entities in an agent-based simulation. They represent individuals, entities, or elements in the system being modeled. Each agent has its own set of attributes, behaviors, and decision-making processes.

\textbf{Environment.} The environment is the space in which agents operate and interact. It includes the physical space, as well as any external factors, e.g., weather conditions, economic changes, political shifts, and natural disasters, that influence agent behavior. Agents may be constrained or influenced by the environment, and their interactions can have effects on the environment itself.

\textbf{Interaction.} Agents interact with each other and their environment through predefined mechanisms. Interactions can be direct (agent-to-agent) or indirect (agent-to-environment or environment-to-agent).

With the above components, agent-based modeling and simulation provide a bottom-up perspective to study the macro-level phenomenons and dynamics from the individual interactions.

\subsubsection{Agent capability}
To achieve realistic simulation in a wide range of application domains, agents should have the following capabilities in terms of perception, decision and action \cite{wooldridge1995intelligent}:
	
	\textbf{Autonomy.} Agents should be able to operate without the direct intervention of humans or others, which is important in real-world applications such as microscopic traffic flow simulation \cite{lopez2018microscopic} and pedestrian movement simulation \cite{batty2003agent}.
	
	\textbf{Social ability.} Agents should be able to interact with other agents (and possibly humans) to complete the assigned goals. When studying social phenomena, group behavior, or social structures, the sociability of agents is key. This includes simulating the formation of social networks, the dynamics of opinions, the spread of culture, and more. The social interactions between agents can be either cooperative or competitive, which are critical when simulating economic activities such as market behavior, consumer decisions, etc.
	
	\textbf{Reactivity.} Agents should be able to perceive their environment and respond quickly to changes in the environment. This capability is especially important in systems that need to simulate real-time responses, such as traffic control systems and automated production lines, and in disaster response scenarios where agents need to be able to respond to environmental changes immediately to effectively conduct early warning and evacuation. More importantly, agents should be able to learn from previous experience and adaptively improve their responses, similar to the idea of reinforcement learning \cite{lin1992self}.
	
	\textbf{Pro-activeness.} Agents should be able to exhibit goal-directed behavior by taking the initiative instead of just responding to their environment. For example, agents need to proactively provide help, advice, and information in applications such as intelligent assistants and actively explore their environment, plan paths, and perform tasks in fields such as autonomous robots and self-driving cars.
	
	It is worth mentioning that, like humans, agents cannot make perfectly rational choices due to limitations of knowledge and computational capacity \cite{simon1997models}. Instead, they can make suboptimal yet acceptable decisions based on imperfect information. This capability is particularly critical in achieving human-like simulations in the economic market \cite{arthur1991designing} and management organizations \cite{puranam2015modelling}. For example, considering agents' bounded rationality when simulating consumer behavior, market transactions, and business decisions can more accurately reflect real economic activities. In addition, in simulating decision-making, teamwork, and leadership within organizations, bounded rationality helps reveal behavioral dynamics in real work settings.

\subsubsection{Applications of agent-based modeling and simulation}

The flexibility of agent-based modeling and simulation allows for the exploration of diverse scenarios and the study of emergent phenomena in a controlled simulation environment. Therefore, it offers researchers and practitioners a versatile tool for understanding and predicting the behavior of complex systems across various domains. 

Based on the four categories of the target systems, current applications of agent-based simulation can be divided into \textbf{four domains}:

\textbf{Physical domain.} This category refers to the natural system in the physical environment \cite{an2012modeling}. Typical applications include \textbf{ecology and biology~\cite{zhang2020overview,pereira2004agent}.}, such as modeling ecological systems~\cite{heckbert2010agent,lippe2019using}, species interactions~\cite{mclane2011role}, and the impact of environmental changes~\cite{pertoldi2004impact,beltran2017agent}.
Many simulation problems in urban environments also belong to the physical domain~\cite{an2012modeling}, such as transportation, human mobility, etc.
Specifically, for urban planning~\cite{gaube2013impact}, agent-based modeling and simulation can aid in simulating urban growth~\cite{arsanjani2013spatiotemporal,barros2004urban}, traffic patterns~\cite{mastio2018distributed,de2019mesoscopic}, and the impact of urban policies~\cite{maggi2016understanding,widener2013agent,ma2013agent}. 
Another application is \textbf{engineering and manufacturing~\cite{barbosa2011simulation,rolon2012agent}.}, in which agent-based molding and simulation can be applied to model supply chain dynamics~\cite{schieritz2003emergent}, production processes~\cite{parv2019agent}, and the interactions of entities within manufacturing systems. 

\textbf{Social domain.} The social domain mainly covers the social behavior simulation, which can be further divided into 1) social interaction that focuses on social networks, community interactions, or organizational behavior \cite{macy2002factors,wall2016agent}, and 2) economic system that simulates economic systems, market dynamics, or financial interactions \cite{samanidou2007agent}. Specifically, 
for social sciences~\cite{conte2014agent,gilbert2007computational,gilbert2000build,terna1998simulation}, agent-based modeling and simulation is widely used to model social phenomena such as crowd behavior~\cite{luo2008agent,kountouriotis2014agent}, opinion dynamics~\cite{li2020opinion,banisch2012agent}, and social network interactions~\cite{madey2003agent,el2012social,gilbert2004agent}. The agent-based modeling can simulate the emergence of societal patterns and trends~\cite{helbing2012social}.
As for the research of  economics~\cite{leombruni2005economists,van2008agent,hamill2015agent}, agent-based models are employed to study economic systems~\cite{deguchi2011economics}, market dynamics~\cite{rouchier2017agent,wang2018agent}, and the behavior of individual economic agents~\cite{mueller2016economic}.

\textbf{Cyber domain.} Besides the physical world and human society, our daily life has been further extended into cyberspace. Therefore, agent-based simulation has also been applied in wide areas like web-based behaviors \cite{guyot2006agent} and cyber-security applications \cite{alluhaybi2019survey}.

\textbf{Hybrid domain.} This category includes hybrid systems combining components covering the physical world, social life, and cyberspace. For example, an urban environment is a socio-physical environment that integrates social behavior with physical infrastructure. Moreover, it is also multi-layered after taking online social networks into account. That is, these applications involve more than one domain of physical, social, or cyber domains.
Therefore, agent-based simulations within an urban environment, such as urban planning \cite{chen2012agent} and epidemic control \cite{silva2020covid}, are far more complex and challenging than those in unitary environments.
Moreover, for \textbf{healthcare~\cite{cabrera2011optimization,barnes2013applications}}, agent-based modeling and simulation can be used to model the spread of infectious diseases~\cite{perez2009agent}, healthcare systems~\cite{silverman2015systems}, and the effectiveness of interventions~\cite{beheshti2017comparing}, which help in understanding and planning for public health scenarios.

\subsubsection{Methodologies of agent-based modeling and simulation}
The development of modeling technologies utilized in agent-based simulation has also gone through the early stage of knowledge-driven approaches and the recent stage of data-driven approaches. Specifically, the former includes various approaches based on predefined rules or symbolic equations, and the latter includes stochastic models and machine learning models.
\begin{itemize}[leftmargin=*]
\item \textbf{Predefined rules.} This approach involves defining explicit rules that govern agent behaviors. These rules are typically based on logical or conditional statements that dictate how agents react to specific situations or inputs. The most well-known example is the cellular automata \cite{wolfram1984cellular} that leverages simple, local rules to simulate complex global phenomena that exist not only in the natural world but also in complex urban systems.

\item \textbf{Symbolic equations.} Compared with predefined rules, symbolic equations are used to represent relationships or behaviors in a more formal, mathematical manner. These can include algebraic equations, differential equations, or other mathematical formulations. A typical example is the social force model widely used in pedestrian movement simulation \cite{helbing1995social}. It assumes that pedestrian movements are driven by a Newton-like law decided by an attractive force driven by the destination and a repulsive force from neighboring pedestrians or obstacles. 

\item \textbf{Stochastic modeling.} This approach introduces randomness and probability into agent decision-making, which is useful for capturing the uncertainty and variability inherent in many real-world systems \cite{feng2012linking}. For example, to account for the impact of randomness originating from human decision-making, we can leverage discrete choice models for simulating pedestrian walking behaviors \cite{antonini2006discrete}.

\item \textbf{Machine learning models.} Machine learning models allow agents to learn from data or through interaction with their environment. 
Supervised learning approaches are generally used for estimating parameters of agent-based models, while reinforcement learning approaches are widely used in the simulation period, enhancing the adaptation capability of agents within dynamic environments \cite{kavak2018big, kim2021automatic, platas2023survey}.
\end{itemize}

\subsubsection{Limitations}
Early works on agent-based simulation are keen to design ``deliberative architectures'' that rely on explicit, often complex, internal models to make decisions, emphasizing the importance of planning, reasoning, and decision-making processes \cite{wooldridge1995intelligent}. However, optimizing the internal world model and planning-reasoning module based on symbolical AI approaches are generally intractable in practice. This leads to the prevalence of ``reactive architectures'' in agent-based simulations, which instead rely primarily on direct sense-action loops rather than complex internal models of the world or deep reasoning processes to make decisions. The subsequent development of AI, especially deep learning technology, does not fundamentally change this paradigm of agent-based simulation due to the poor interpretability and generalization capability. However, facing the need for realistic simulation of real-world processes or systems, current approaches still have several limitations, as described below.

\textbf{Simple agent architecture is not enough to cope with complex tasks.} Although ``reactive architectures'' are able to adapt to different environmental conditions, they may be limited in handling complex tasks or situations that require long-term planning. To achieve human-like simulation in real-world complex problems, current agent architecture requires redesigns that solve challenges in processing speed, resource efficiency and task complexity. Specifically, agents should be capable of complex planning and reasoning processes, like using internal models to predict the consequences of different courses of action and choose the best one, and able to develop and execute complex strategies to achieve long-term goals.

\textbf{It is difficult to develop a general agent that can support simulations across environments.} Different environments vary in dimensions like complexity, dynamics, and uncertainty. Due to this diversity, a specific agent that is effective in one environment (like a financial market simulation) might be completely ineffective in another (like a social campaign simulation). In real-world applications where the target environment is often hybrid with significant dynamics and uncertainty, developing specific agents case by case is highly inefficient and costly.

\textbf{Existing methods cannot support integrative simulation in real-world problems.} A versatile agent-based simulation model should be able to describe how systems operate under known conditions, explain why certain patterns emerge, predict future states based on existing observations, and explore the outcomes of hypothetical scenarios. However, existing methods cannot support the above tasks simultaneously: rule-based methods are useful in descriptive problems, while symbolic or stochastic methods can provide explanations regarding underlying mechanisms that drive the system. Comparatively, machine learning models are better at predictive problems by learning hidden patterns from data but with less interpretability. Therefore, there remain challenges in developing methods that simultaneously capture the accuracy of behavioral modeling, interpretability of mechanisms, adaptability, and reliability under environmental changes.

\subsection{Large language models and LLM-empowered agents} 

Large language models (LLMs), such as ChatGPT~\cite{chatgpt}, Gemini~\cite{gemini}, LLaMA~\cite{touvron2023llama}, Alpaca~\cite{alpaca}, and GLM~\cite{zeng2023glm}, are the latest paradigm of language models, which evolve from early statistical language models~\cite{bellegarda2004statistical} to neural language models ~\cite{melis2017state}, then to pre-trained language models~\cite{brown2020language}, and finally to large language models~\cite{zhao2023survey}.
With billions of parameters and extensive pre-training corpus, LLMs have shown astonishing abilities not only in natural language processing tasks~\cite{li2023seed,zhang2023benchmarking} such as text generation, summarization, translation, etc., but also in complex reasoning and planning tasks, such as solving mathematical problems~\cite{arora2023have}, etc.
Pre-training on large-scale corpora lays the foundation ability for zero-shot generalization. Moreover, pre-trained models can be further fine-tuned for specific tasks, adapting to particular application scenarios~\cite{jiang2023health}. 
In addition, the advances of large language models in the past year such as ChatGPT and GPT-4 have achieved human-like reasoning ability, a milestone that is now being considered to be the seed of artificial general intelligence (AGI).
Specifically, the capacity to acquire and use language is a key aspect of how we, humans, distinguish ourselves from other beings~\cite{tomasello2010origins}.
Language is one of the most important mechanisms we have to interact with the environment, and language provides the basis for high-level abilities~\cite{hauser2002faculty}.

Thus, it is promising to construct large language model-empowered agents~\cite{wang2023survey,xi2023rise} due to their human-like intelligence in perceiving the environment and making decisions.
First, the LLM agent is able to adaptively react and perform tasks based on the environment without predefined explicit instructions~\cite{autogpt, babyagi}. In addition, during the simulation process, the LLM agent can even form new ideas, solutions, goals, etc,~\cite{franceschelli2023creativity}. For example,
AutoGPT~\cite{autogpt} can automatically schedule plans when given a set of available tools and the final task goal, exemplifying the significant potential of LLMs in constructing autonomous agents. Meanwhile, BabyAGI~\cite{babyagi} created an LLM-driven script running an infinite loop, which continuously maintains a task list, in which each task is completed the task by ChatGPT API~\cite{chatgpt} based on the task context.
Second, the LLM agent has enough intelligence that it can respond like a human and even actively take actions with self-oriented planning and scheduling~\cite{wang2023survey,xi2023rise}. The environment input is not limited to text; rather, recent multi-modal fusion models can be fed other types of information, such as image or audio~\cite{zhu2023minigpt}. The action space of the LLM agent is neither limited to text, for which the tool-usage ability allows the agent to take more actions~\cite{schick2023toolformer}.
Lastly, the LLM agent has the ability to interact and communicate with humans or other AI agents~\cite{park2023generative}. In the simulation, especially agent-based simulation, the agent's communication ability elevates individual simulation to the community level~\cite{gilbert2005simulation}. An LLM-driven agent can generate text, which can be received and understood by another agent, in turn providing the basis for interpretable communication among agents or between humans and agents~\cite{park2023generative}. Moreover, the simulation at the community level requires heterogeneity of agents, and the LLM agents can meet these requirements for playing different roles in society~\cite{qian2023communicative}.
An artificial society constructed by LLM agents can further reveal the emergence of swarm intelligence with collective agent behaviors~\cite{gao2023s,park2023generative}, similar to wisdom-of-crowds in human society~\cite{surowiecki2005wisdom}.

As mentioned above, the simulation system has widely utilized the paradigm of agent-based modeling, which requires agents with high-level abilities, well motivating the use of large language model-empowered agents in simulation scenarios.

\section{Critical abilities of LLM for agent-based modeling and simulation} 

\begin{figure}[t!]
	\centering
	\includegraphics[width=0.97\textwidth]{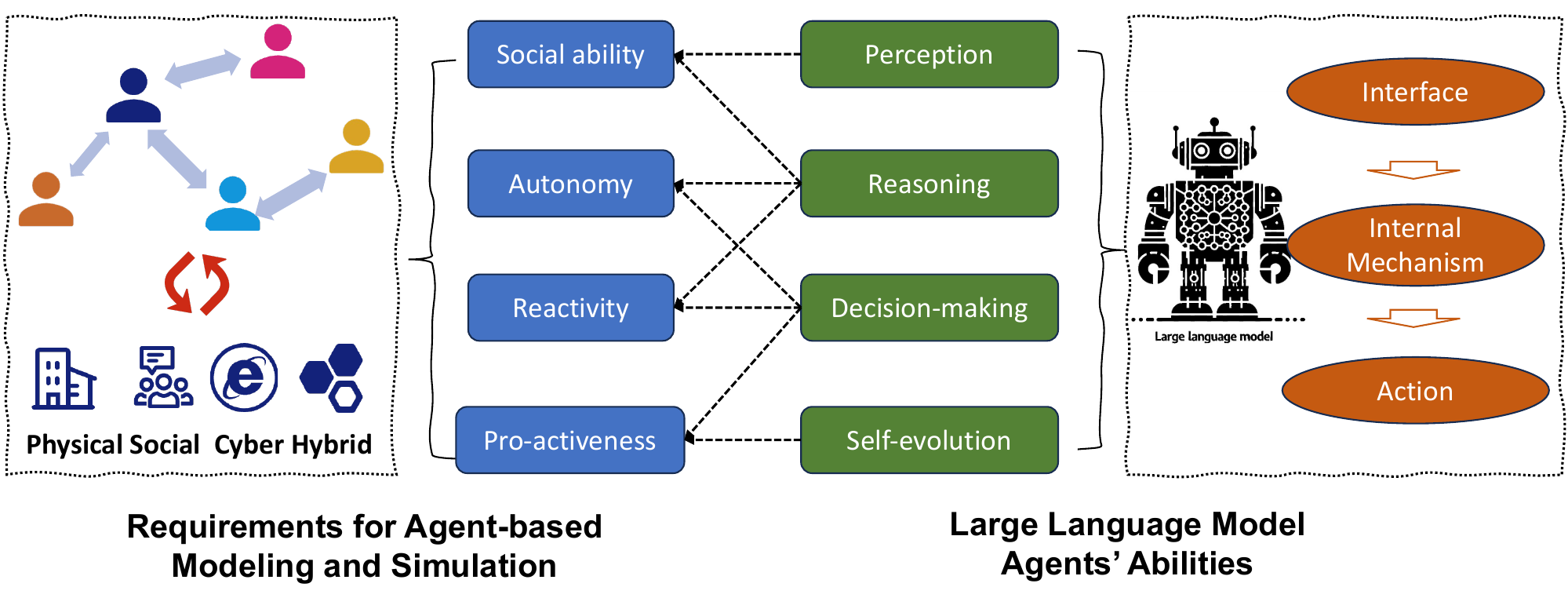}
	\caption{Illustration of how large language model agents meet the requirements of agent-based modeling and simulation.}
	\label{fig:abs_lmm}
\end{figure}

As mentioned above, agent-based modeling and simulation serve as a basic approach for simulation in many areas~\cite{macal2005tutorial,elsenbroich2014agent}, but it still suffers from several key challenges.
Large language model-empowered agents not only meet the requirements for agent-based simulation but also address the limitations relying on their strong abilities in perception, reasoning, decision-making, and self-evolution, illustrated in Figure~\ref{fig:abs_lmm}.
% XXXX\todo{complete it.}

\subsection{Perception} 
The core of agent-based modeling and simulation is to model how an individual agent interacts with an environment~\cite{macal2005tutorial}, which requires the agent to accurately sense various types of information from said environment.
As for the large language model-empowered agents, 
the ability of language enables agents to comprehend and respond to diverse environments directly or indirectly.
On the one hand, the basic ability to understand and generate text enable agents to engage in complex dialogues, negotiate, and exchange information, and support direct interaction.
On the other hand, the interface between the agent and environment can be operated via texts~\cite{xagent2023}, which leads to indirect interaction.
Of course, such ability also supports the communication between different agents, besides the agent-environment perspective.

It is worth mentioning that the ability to interact with the environment and other agents is not adequate to achieve human-like simulations.
To be more specific, it is also required that large language model-based agents ``put themselves in real humans' shoes", thereby allowing the agent imagine that it is \textit{indeed in the environment}.
That is, LLM agents should be able to comprehend, perceive, and respond to diverse needs, emotions, and attitudes within different contexts, from the ``first-view sight''~\cite{shanahan2023role}. This capability enables models to better understand the information from the environment or other agents and generate more real responses. 

\subsection{Reasoning and decision making}
One critical challenge in traditional agent-based simulation is that rule-based or even neural network-based agent is \textit{not intelligent enough}~\cite{cipi2011simulation}. That is, the agent is not able to make correct or optimal decisions, such as choosing a crowded road in transportation simulation or sending an incorrect message in social network simulation.
This can be explained by the fact that the traditional neural network-based artificial intelligence is still not as intelligent as a real human~\cite{hoshen2017iq,liu2019well,mandziuk2019deepiq,hernandez2016computer}. In contrast, large language model-empowered agents exhibit heightened reasoning capabilities, enabling them to make more informed decisions and choose suitable actions within the simulation. 
% Another dimension of intelligence is about whether it can sense the input and intervention, corresponding to the ability to comprehend varied inputs, including textual instructions, user queries, or environmental cues, fostering a more intelligent agent.
Despite making suitable decisions, another critical advantage of large language model-empowered agents to support better agent-based modeling and simulation is autonomy~\cite{fu2023drive}. 
With only limited guidance, regulations, and goals, agents equipped with large language models can autonomously take actions, make plans for the given goal, or even achieve new goals without the need for explicit programming or predefined rules~\cite{park2023generative}. 
That is, autonomy enables LLM agents to dynamically adjust their actions and strategies based on real circumstances, contributing to the realism of the simulation.

\subsection{Adaptive learning and evolution}

For agent-based modeling and simulation, the system always has uncertainty and uncontrollability~\cite{macal2005tutorial}.
In other words, the environment and the agent's state may be completely different compared with the initial stage of the simulation. 
As the old story of \textit{Rip Van Winkle} tells, a man falls asleep in the mountains and awakens to find that the world around him has drastically changed during his slumber.
That is, the environment is continuously changing in a long-term social network simulation~\cite{gao2023s}; the agent should be able to adapt to the new environment, formulating decision policies that may deviate significantly from their original strategies.
Obviously, adaptive learning and evolution are challenging for traditional approaches, but luckily, this can be addressed by large language model-based agents~\cite{lu2023self}.
% To conclude, the integration of large language models enables agents in simulations to engage in adaptive learning and evolutionary processes. 
Specifically, with the ability to continually learn from new data and adapt to changing contexts, LLM agents can evolve behaviors and decision-making strategies over time. Agents can assimilate new information, analyze emerging patterns in data, and modify their responses or actions accordingly based on in-context learning~\cite{dong2022survey}, mirroring the dynamic nature of real-world entities. This adaptability contributes to the simulation's realism by simulating the learning curve and evolution of agents' behaviors in response to varying stimuli.

\subsection{Heterogeneity and personalizing}

As the saying goes, \textit{one man's meat is another man's poison}.
Heterogeneity of agents is critical for agent-based simulation, with the complex society~\cite{brown2006effects} or economic system~\cite{bohlmann2010effects} with heterogeneous individuals. 
Specifically, in agent-based modeling and simulations, the heterogeneity of agents involves representing diverse characteristics, behaviors, and decision-making processes among individuals. Agent-based simulation stands out for its capacity to accommodate varied rules or parameters compared to traditional simulation methods, discussed as follows.

The first one is the extremely high complexity of parameter settings of the existing methods~\cite{elliott2002exploring,macal2005tutorial}. In these models, the vast array of variables influencing an agent's behavior—from personal traits to environmental factors—makes selecting and calibrating these parameters daunting. This complexity often leads to oversimplification, compromising the simulation's accuracy in portraying true heterogeneity~\cite{macal2005tutorial}. Moreover, acquiring accurate and comprehensive data to inform parameter selection is another challenge. That is, real-world data capturing diverse individual behaviors across various contexts might be limited or challenging to collect. Furthermore, validating the chosen parameters against real-world observations to ensure their reliability adds another layer of complexity.

Second, the rule or the model cannot cover all dimensions of heterogeneity, as real-world individuals are very complex~\cite{macal2005tutorial}. Using rules to drive agent behaviors only captures certain aspects of heterogeneity but could lack the depth to encapsulate the full spectrum of diverse behaviors, preferences, and decision-making processes. 
Furthermore, as the model capacity, trying to cover all dimensions of heterogeneity within a single model is too idealistic.
Thus, balancing model simplicity and accurate modeling agents becomes a critical challenge in agent-based modeling and simulation, resulting in oversimplification or neglect of certain aspects of agent heterogeneity.

Different from the traditional methods, the LLM-based agents support 1) capturing complex internal characteristics with internal human-like cognitive complexity, and 2) specialized and customized characteristics with prompting, in-context learning, or fine-tuning. 

% \clearpage
\section{Challenges and approaches of LLM agent-based modeling and simulation}
% \todo{cite our paper}
% \task{Gao Chen, Li Nian, Lan Xiaochong}  3.5page}
The core of agent-based modeling and simulation is how the agent reacts to the environment and how agents interact with each other, in which agents should behave close to real-world individuals with human knowledge and rules, as real as possible.
Therefore, when constructing large language model-empowered agents for simulation, there are four major challenges, including perceiving the environment, aligning with human knowledge and rules, choosing suitable actions, and evaluating the simulation.
% \todo{use proper sentences to lead to these four conclusions.}
We will discuss the solutions from a high-level perspective, and how the existing works address them will be elaborated on in detail in the next section.

\subsection{Environment construction and interface}
% \task{Gao Chen} 1 page} 

For agent-based simulation with large language models, the first step is to construct the environment, virtual or real, and then design how the agent interacts with the environment and other agents. Thus, we need to propose proper methods for an environment that LLM can perceive and interact with. 
% How to build the simulation environment, how to make large models perceive the environment, and how the 

\subsubsection{Environment: define the world and rules}
The external environment in agent-based simulation varies for different domains.
% , including the virtual online environment, socioeconomic environment, etc. 
In general, the environment built by existing works can be divided into two categories: virtual and real.
\begin{itemize}[leftmargin=*]
\item The virtual environment includes simulation applications with predefined rules in prototype-level simulation, such as a virtual social system, game, etc.
For example, Qian~\textit{et al.}~\cite{qian2023communicative} designed a virtual software company with multiple agents for different roles, such as CEO, managers, programmers, etc.
% Gao~\textit{et al.}~\cite{gao2023s} designed a virtual online social network in which agents can post or propagate content.  
Wang~\textit{et al.}~\cite{wang2023recagent} constructed an environment of a virtual recommender system in which agents can browse the recommended contents and provide feedback.
The sandbox environment is one kind of virtual environment where the principles and ideas conceptualized in a virtual environment can be tested and adapted to real-world applications. For example, Generative Agent~\cite{park2023generative} builds a Smallville sandbox world in which large language model-empowered agents plan their days, share news, form relationships, and coordinate group activities.

\item The real environment includes our real world. 
For example, Li~\textit{et al.}~\cite{li2023large} deploy large language model-based agents to simulate the economic activities, in which agents can represent both consumers and workers.
WebAgent~\cite{gur2023real} simulates a real human browsing and accessing online content of real websites.
UGI~\cite{ugi} proposed to build agents for the real-world urban environment, and the agents are expected to generate various human behaviors in the city, including navigation, social, economic, etc.
% Schumann~\textit{et al.}~\cite{schumann2023velma} proposed 
% VELMA is an embodied language model agent that integrates verbalized trajectory instructions and visual observations to guide its actions. VELMA successfully follows navigation instructions in Street View with minimal examples for accurately simulating human mobility behaviors in the real world. 

\end{itemize}

\subsubsection{Interface}
The interface actually has two aspects, how the agent interacts with the environment and how agents communicate with each other.

\begin{itemize}[leftmargin=*]
\item Input and output of the environment. Most existing works use texts as the major interface, naturally due to the ability to understand and generate texts of large language models. Even if the environment is a sandbox with rich models, such as Smallville sandbox world~\cite{park2023generative} in which the environment is still represented with texts, relying on which the agent can perceive the context. 
In addition, the basic rules or domain-specific knowledge, such as the game rule, is also summarized with texts, which are received by the large language model agents with prompt engineering.
Due to the limitation of texts, the existing works construct various tools to interact with complex environments or data, but recalling or using these tools is still based on texts. For example, in ~\cite{zhu2023ghost}, the large language model's action is a phrase, which can be a parameter of the tool function to interact with the simulation environment.

\item Communication between agents. First, the direct communication between agents is also focusing on the texts. For example, in agent-based simulation for social science, the textual information exchange represents the communication between humans in the real world. Second, the agents can indirectly interact with others through predefined rules; for example, in economic simulation, the agents can work in the same factory, and the rule in the economic system makes them interact indirectly.

\end{itemize}

% \begin{enumerate}
%     \item Text
%     \item Engineering prompts
%     \item Tool interfaces
% \end{enumerate}

% \clearpage
% \subsection{Human alignment. How to align with human knowledge and values; how to achieve personalized simulations for different individuals: Engineering prompts, fine-tuning, and prompt fine-tuning \task{Li Nian} 1 page}
\subsection{Human alignment and personalization}
% \task{Li Nian}}

% \begin{itemize}
%     \item AlpacaFarm: A Simulation Framework for Methods that Learn from Human Feedback
%     \item Role play with large language models
% \end{itemize}

Although LLMs have already demonstrated remarkable human-like characteristics in many aspects, agents based on LLMs still lack the necessary domain knowledge in specific areas, leading to irrational decisions. Therefore, aligning LLM agents with human knowledge and values, especially those of domain experts, is an essential challenge to achieve more realistic domain simulations. However, the heterogeneity of agents, as a fundamental characteristic of ABM, is both an advantage and a challenge for traditional models. While, LLMs possess a powerful capability to simulate heterogeneous agents, ensuring controllable heterogeneity. However, enabling LLMs to play different roles to meet personalized simulation requirements is a significant challenge. Next, we will explain the methods and technologies to address these two challenges from two perspectives: prompt engineering and tuning, and introduce the existing related work in these areas.

\subsubsection{Human alignment}

\textbf{Prompt engineering}.
When simulating specific agents, we can provide task instructions, background knowledge, generation patterns, and task examples specific to certain domains or scenarios, thereby aligning LLMs' output with human knowledge and values when deployed. For example, providing detailed descriptions of game rules and examples for the agent allows it to consider various factors it cares about, like humans when making decisions, such as self-interests, fairness, etc~\cite{akata2023playing}. In addition, constructing modules such as reflection and memory can improve agents' planning and reasoning capabilities, thereby giving them stronger gaming capabilities and creating a possible path towards human-intelligent gaming~\cite{guo2023suspicion}.

\textbf{Tuning}.
Tuning requires constructing a training dataset for specific domains, scenarios, or hiring domain experts. Based on the dataset or expert feedback, fine-tuning the LLM can also empower the agents with more domain-specific knowledge, producing outputs more in line with human knowledge and values. For example, Singhal \textit{et al.}~\cite{singhal2023large} propose to achieve knowledge alignment in clinical medicine. The proposed MultiMedQA benchmark combines six existing medical question-and-answer datasets covering professional medicine, research, and consumer inquiries. Additionally, Med-PaLM~\cite{singhal2023large}, a LLM for the medical field, is trained based on a foundational model PaLM~\cite{chowdhery2023palm}. In terms of implementation, the authors incorporate examples of medical question-and-answer and modify model prompts through the guidance of professional clinicians (involving five clinical doctors) for fine-tuning. This guides the model to generate text consistent with clinical requirements. With this domain-specific LLM, we can simulate agents~(\textit{e.g.},  medical assistants) in real-world medical environments. In addition to collecting large-scale datasets with domain knowledge, other research~\cite{dubois2023alpacafarm} directly uses LLMs to generate  ``human feedback'', specifically pair-wise feedback for instructions, for LLM fine-tuning. Results show that the generated feedback enables LLM to achieve high human alignment 45$\times$ cheaper than hiring crowd workers to give feedback in experiments.

\subsubsection{Personalization}

\textbf{Prompt engineering}.
The basic idea is to adapt to personalized needs by providing LLM agents with individual preferences, expected output patterns, background knowledge, etc., thereby making the output closer to the specific needs or preferences of individuals when deployed. For example, in the well-known LLM-based social activity simulation, AI Town~\cite{park2023generative}, personalized interaction behaviors of agents in different scenarios, at different times, and with different other agents can be achieved by introducing professions, behavioral preferences, and interpersonal relationships in the prompts. In economic simulation, specifically simulation of canonical games, the agent's preferences can be specified in the prompt, such as cooperative, selfish, altruistic, etc., so that the agent will have different levels of cooperative tendencies during the game playing~\cite{phelps2023investigating}.

\textbf{Tuning}.
Tuning for personalization requires selectively constructing datasets or fine-tuning multiple models based on feedback from different users, with each model corresponding to one or a type of personalized needs. This can also be achieved by using specific combinations to provide relevant, personalized requirements. Some research attempts to efficiently align LLMs with various preferences tailored to different users' distinct preferences~\cite{jang2023personalized}. Specifically, user preferences are decomposed into standards across multiple aspects, with personalized optimization based on RLHF targeted towards different aspects. In practical applications, the strategy of LLM response generation is based on linearly weighting strategies according to user preferences. When simulating agents with individual preferences~(\textit{e.g.}, users in recommender systems), this approach achieves a more accurate match for different preferences and is also easily generalizable to scenarios with a broader range of preferences.

% \clearpage
\subsection{How to simulate actions}
% \task{Lan Xiaochong} 1 page}
This section aims to delve into how LLM agents are designed to exhibit complex behaviors that are reflective of real-world cognitive processes. This involves understanding and implementing the mechanisms by which these artificial agents can retain and utilize past experiences (memory)~\cite{park2023generative,gao2023s,zhu2023ghost}, introspect and adjust their behavior based on their outcomes (reflection)~\cite{park2023generative,shinn2023reflexion}, and execute a sequence of interconnected tasks that mimic human workflows (planning)~\cite{wei2022chain}.

\subsubsection{\textbf{Planning}}
Here, we introduce the methodology by which LLM agents approach complex tasks through decomposition. Initially, an LLM assesses the task to understand its main objectives and context. It then breaks down the task into smaller, manageable subtasks, each contributing towards the overall goal. This segmentation leverages the LLM's training corpus to recognize patterns and apply relevant knowledge efficiently~\cite{zhu2023ghost,wang2023voyager,sun2023adaplanner,park2023generative}.

Each subtask is executed sequentially, with the LLM agent applying its knowledge base to ensure logical progression and coherence. This approach not only simplifies complex tasks but also enhances the LLM’s accuracy and adaptability. By tackling tasks incrementally, the LLM agent can adapt its strategies and ensure that each step is contextually relevant and logically structured. 
For example, GITM~\cite{zhu2023ghost} showcases an LLM agent that decomposes the overarching goal of ``Mining Diamond" into a series of sub-goals, constructing a sub-goal tree. This model uses its text-based knowledge and memory to navigate in a virtual environment, making strategic decisions at each tree node to achieve the main objective. Voyager~\cite{wang2023voyager} employs an automatic curriculum to aid the LLM agent in understanding the sequence of actions required to reach a goal. By reasoning with the available resources, the LLM agent can plan an efficient course of action, like upgrading tools for better efficacy and demonstrating adaptive problem-solving skills. AdaPlanner~\cite{sun2023adaplanner} introduces an LLM that refines its action plan based on feedback, which has an in-plan refiner for aligning actions with predictions and an out-of-plan refiner to adjust when predictions don't match outcomes, showcasing the model's ability to adapt and revise its plan dynamically in response to changing scenarios.

In summary, advancements represent significant strides in task decomposition and strategic planning. They highlight the capability of LLMs to not only break down complex tasks into manageable sub-goals but also to dynamically adapt their strategies and refine plans based on ongoing feedback and changing scenarios, thereby enhancing decision-making and problem-solving efficiency in various contexts.
\subsubsection{\textbf{Memory}}

Human behavior is largely influenced by past experiences and insights, which are stored in memory. If LLM agents aim to mimic this aspect of human behavior, they also need to reference past experiences and insights when acting. However, the volume of this information is often immense, frequently exceeding the context window length of LLMs. Therefore, it's necessary to design a memory system that functions as an external database for LLM agents. This system should have appropriate mechanisms for organizing, updating, and retrieving information, enabling LLM agents to reference these memories for future actions.

Generative Agents~\cite{park2023generative} showcases an LLM agent that develops a generative memory system, integrating sensory perceptions with a continuous stream of experiences. This system not only stores information but actively engages in planning and reflection, adapting its behavior based on past outcomes. Chen et al.~\cite{chen2023put} illustrate an LLM agent's strategic prowess in auction scenarios, where it adapts its bidding strategy by synthesizing new information with existing memories to maximize profits or meet specific objectives.

GITM~\cite{zhu2023ghost} and Voyager~\cite{wang2023voyager} are seen curating a skill library, updating its capabilities through practice and feedback. This approach reflects an understanding of task requirements and environmental challenges, where the LLM's memory serves as a dynamic repository of actions and strategies. The distinction between explicit and implicit memory comes into play in simulations that require the LLM to navigate complex tasks, such as resource management and goal-oriented action planning in open-world environments. Here, the LLM's memory functions extend beyond simple recall, enabling the agent to perform with a sense of history and progression.

Lastly, the role of memory in social interactions is explored through simulations in S3~\cite{gao2023s} that mimic the intricacies of human behavior. LLMs track and adapt to changing social cues and demographic shifts, employing memory not just as a record of past interactions but as a tool for future social navigation and decision-making. Li et al.~\cite{li2023large} demonstrate how a memory module in LLMs can be crucial for understanding and adapting to dynamic social environments. They show that LLMs, equipped with a memory of past social interactions and trends, can more effectively predict and respond to future economic changes, enhancing their decision-making in complex social landscapes. 

Collectively, these studies contribute to our understanding of LLMs as agents capable of sophisticated memory management, crucial for their function in dynamic and unpredictable environments. They highlight the remarkable potential of LLMs to transcend traditional data storage, moving towards a more integrated and intelligent use of memory in artificial cognition.

\subsubsection{\textbf{Reflection}}

The section explores how LLM agents incorporate feedback mechanisms to enhance their memory systems, improving decision-making and learning processes. This reflection encompasses both short-term and long-term memory facets, enabling LLMs to adapt their behaviors and strategies dynamically.

An exemplary implementation of this reflective cycle is Reflexion~\cite{shinn2023reflexion}. In this work, the LLM leverages an integrated evaluator to internally assess the efficacy of actions based on the rewards received. It also utilizes a prompt-based approach to self-reflection, allowing the agent to internally simulate and critique its performance. This dual feedback system enables the agent to refine its memory and behavior in a nuanced and continuous learning process. The model captures short-term memory as trajectories of actions and observations, while long-term memory encompasses accumulated experiences. The interaction between these memory types and the reflective loop ensures that the agent's memory is not only a repository of past events but also a dynamic foundation for future improvement and learning. This system exemplifies how LLMs can evolve from static knowledge bases to dynamic entities capable of self-improvement through iterative reflection and adaptation. In S3~\cite{gao2023s}, the LLMs' ability to reflect is intricately tied to their simulation of human social interactions, where they continuously adjust their understanding and responses based on evolving social dynamics and cues. This reflective capacity enables them to navigate complex social environments with greater finesse. In the work of Li et al.~\cite{li2023large}, reflection is leveraged to refine the LLMs' approach to socio-economic predictions. By reflecting on past interactions and trends, these models can adapt their predictive algorithms, leading to more accurate responses to future social and economic shifts.

In summary, in the realm of simulating actions, LLMs stand out for their ability to integrate planning, memory, and reflection. They employ a cyclical approach where planning dictates the course of action, memory provides a knowledge base derived from past experiences, and reflection adjusts strategies based on feedback. This dynamic interplay allows LLMs to not only execute actions within varied simulations but also to continuously learn and adapt. By simulating these cognitive processes, LLMs demonstrate an advanced capacity for autonomous decision-making, which is increasingly indistinguishable from human-like behavior in complexity and adaptability.
% \clearpage

\subsection{Evaluation of LLM agents}
% \subsection{How to evaluate simulations of large language model agents}
% \task{Gao Chen} 0.5page}
\subsubsection{Realness validation with real human data}
The basic evaluation protocol for LLM-based agents is to compare the simulation's output with existing real-world data. 
The evaluation can be conducted at two levels: micro-level and macro-level. Specifically, micro-level evaluation refers to evaluating the ability to simulate the individual agent's behavior or actions as realistically as possible.
For example, in S$^3$~\cite{gao2023s}, the authors test the performance of the LLM agents in predicting the individual agent's next state, given the current state and the environment context.
On the other hand, since the agent-based simulation always pays more attention to the emerged phenomenon of the population, macro-level evaluation is of great significance, which aims to evaluate whether the simulated process has the same pattern, regularity, etc., as the real-world data.
In S$^3$~\cite{gao2023s}, one of the main goals is to accurately predict the dynamics of information, opinion, and attitude based on the collected real-world social media data.
In economics simulation~\cite{li2023large}, the simulated economic system is evaluated on whether can emerge those most representative macroeconomic regularities, such as Okun's law~\cite{plosser1979potential}, etc.
% It is important for many simulation scenarios whether the agent can behave close to a real human. 
Furthermore, the generated behaviors' rationality can also be evaluated, such as logical consistency, adherence to established common sense, or following the given rule in the simulation environment.
% \subsubsection{Testing capabilities on standard tasks.}
In addition, we can assess the simulated agent's performance against established benchmarks or standardized tasks relevant to its domain. For example, whether the agent can reach human-level evaluation scores in a web-browsing or game environment~\cite{chang2023survey}.

\subsubsection{Provide explanations for simulated behaviors}
One of the main advantages of the large language model-based agent against the traditional rule-based or neural network-based agent is its strong ability to engage in interactive conversation and textual reasoning.
Therefore, to evaluate whether the agent has understood the simulation rules well, accurately perceived the environment, made a choice rationally, etc., we can directly obtain explanations from the large language model-based agent.
We can evaluate whether the agent-based simulation is good by analyzing the explanations and comparing them with the human data or a well-established theory or model.
For example, in economic simulation~\cite{li2023large}, the authors query the large language model agent about the reason for economic decision-making, which well explains the simulated actions and behaviors.

\subsubsection{Ethics evaluation} 
Besides the simulation accuracy or explainability of the large language model-empowered agent-based simulation, the ethics issue is also of great importance. 
The first one is bias and fairness, and it is essential to assess the simulation for biases in language, culture, gender, race, or other sensitive attributes to evaluate whether the generated content perpetuates or mitigates societal biases.
Another concern is harmful output detection since the output of the generative artificial intelligence is hard to control compared with traditional approaches.
Thus, the practitioners of the large language model agent-based simulation should scrutinize the simulation's output for potentially harmful or inappropriate content, including hate speech, misinformation, or offensive material.

% \clearpage
\section{Recent advances of LLM agent-based modeling and simulation} 
\begin{figure}[t!]
	\centering
	\includegraphics[width=0.70\textwidth]{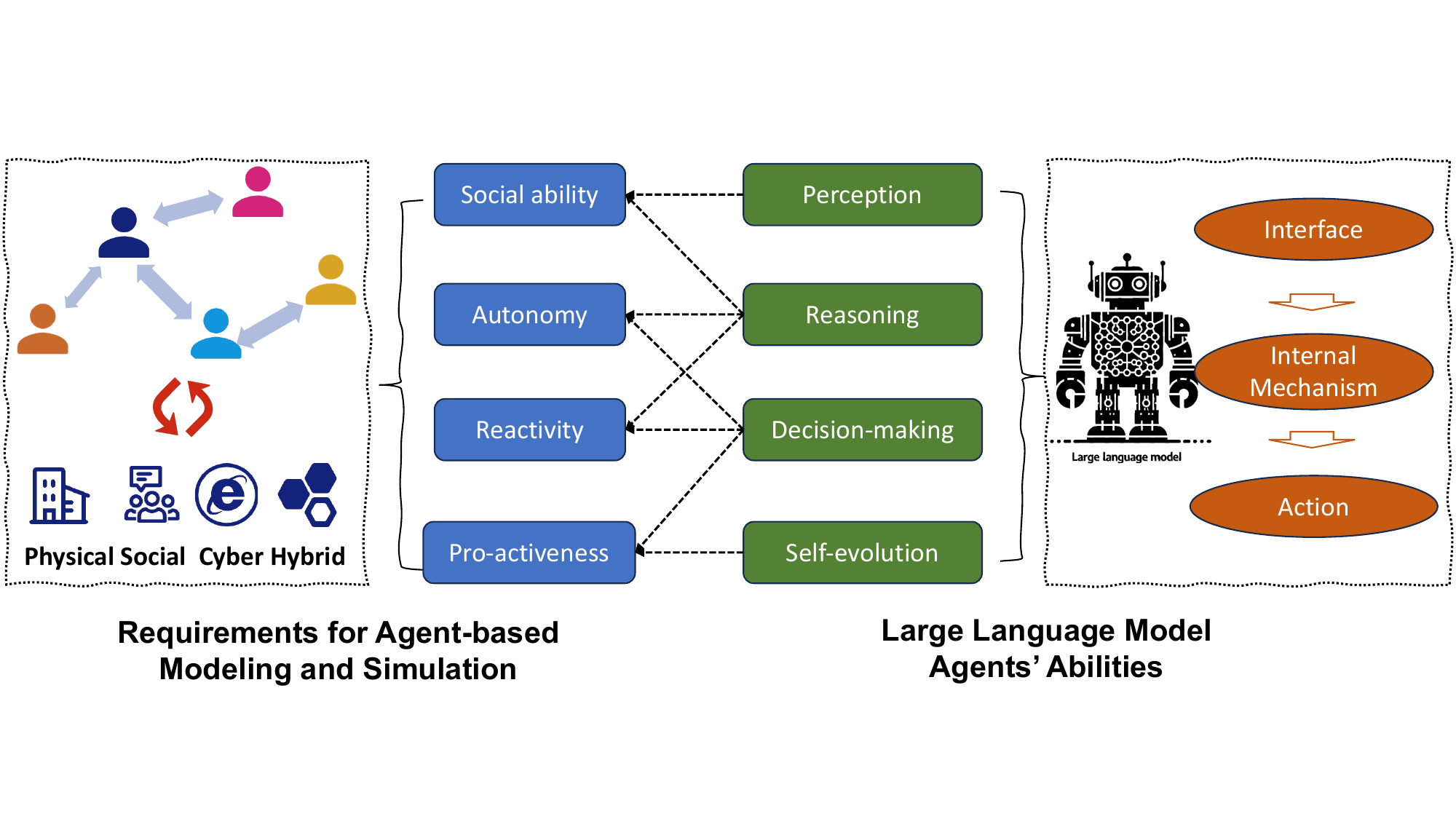}
	\caption{Illustration of LLM agent-based modeling and simulation in different domains.}
	\label{fig:advances}
\end{figure}

In the following, we elaborate on the recent advances to use large language models for agent-based modeling and simulation in social, physical, cyber, and hybrid domains. The typical applications in the first three domains are illustrated in Figure~\ref{fig:advances}, and the details are shown in Table~\ref{tab:advances}.

	\begin{table}[t!]
		\centering
		\footnotesize
		\caption{A list of representative works of agent-based modeling and simulation with large language models.}\label{tab:advances}
		\begin{tabular}{c|c|c|c}
			\hline
			\bf Domain &	\bf  Environment      & \bf Advance        & \bf What to simulate  \\ \hline
			Social & Virtual & Schwitzgebel~\textit{et al.}~\cite{schwitzgebel2023creating}  	& Conversation and interaction  \\ \hline		
			Social	& Virtual             & Xu et al.~\cite{xu2023exploring} & Werewolf game  \\ \hline
			Social	& Virtual            & Acerbi et al.~\cite{acerbi2023large} & Information Propagation \\ \hline
			Social	& Virtual                & Zhang et al.~\cite{zhang2023exploring} & Collaboration Mechanism  \\ \hline
			Social	& Virtual         & Suzuki et al.~\cite{suzuki2023evolutionary} & Cooperation and defection\\ \hline
			Social	& Virtual           & Zarza et al.~\cite{de2023emergent} & Social interaction \\ \hline
			Social	& Real            & Mukobi et al.~\cite{mukobi2023welfare} &Welfare diplomcy game\\ \hline
			Social	& Real      & S3~\cite{gao2023s} & Online social network      \\ \hline
			Social	& Virtual              & SimReddit~\cite{park2022social} & Online forum     \\ \hline
			Social	& Real                & COLA~\cite{lan2023stance} & Cooperative task solving     \\ \hline
			Social	& Virtual             & MAD~\cite{liang2023encouraging} & Cooperative task solving    \\ \hline
			Social	& Virtual       & CHATDEV~\cite{qian2023communicative} & Cooperative task solving   \\ \hline
			Social	& Virtual               & MetaGPT~\cite{hong2023metagpt} & Cooperative task solving      \\ \hline
			Social	& Virtual            & ChatEval~\cite{chan2023chateval} & Cooperative task solving      \\ \hline
			Social	& Virtual                 & CAMEL~\cite{li2023camel} & Cooperative task solving       \\ \hline
			Social	& Virtual             & AgentVerse~\cite{chen2023agentverse} & Cooperative task solving      \\ \hline
			Social	& Virtual             & SPP~\cite{wang2023unleashing} & Cooperative task solving        \\ \hline
			Social	& Virtual       & CoELA~\cite{zhang2023building} & Cooperative task solving   \\ \hline
			Social	& Virtual          &  Humanoid Agents~\cite{wang2023humanoid}                 & Individual social behavior           \\ \hline
			Social	& Real           &  SocioDojo~\cite{liu2023sociodojo}                 & Individual social behavior       \\ \hline
			Social	& Virtual&Liu \textit{et al.}~\cite{liu2023training}           & Individual social behavior           \\ \hline
			Social	& Virtual &Argyle \textit{et al.}~\cite{argyle2023out}                 & Individual social behavior           \\ \hline
			Social	& Virtual &Hamalainen \textit{et al.}~\cite{hamalainen2023evaluating}                 & Individual social behavior           \\ \hline
			Social	& Virtual &Singh \textit{et al.}~\cite{singh2023mind}                 & Individual social behavior           \\ \hline
			Social	& Virtual &Binz \textit{et al.}~\cite{binz2023using}                 & Individual social behavior           \\ \hline
			Social	& Virtual &Elyoseph \textit{et al.}~\cite{elyoseph2023chatgpt}                 & Individual social behavior           \\ \hline
			Social	& Virtual &Li \textit{et al.}~\cite{li2022gpt}                 & Individual social behavior           \\ \hline
			Social	& Virtual       &  Horton~\cite{horton2023large}                  &    Economic system: individual behavior                   \\ \hline
			Social	& Virtual      &  Chen et al.~\cite{chen2023emergence}                 &    Economic system: individual behavior                       \\ \hline
			Social	& Virtual           &  Geerling et al.~\cite{geerling2023chatgpt}                 &    Economic system: individual behavior                             \\ \hline
			Social	& Real           &   Xie et al.~\cite{xie2023wall}                 &    Economic system: market behavior                      \\ \hline
			Social	& Real         &  Faria et al.~\cite{faria2023artificial}                 &    Economic system: market behavior                   \\ \hline
			Social	& Real         &  Bybee et al.~\cite{bybee2023surveying}                 &    Economic system: market behavior                   \\ \hline
			Social	& Virtual         &  Phelps et al.~\cite{phelps2023investigating}                 &  Economic system:  game theory                      \\ \hline
			Social	& Virtual              &  Akata et al.~\cite{akata2023playing}                 &    Economic system: game theory                      \\ \hline
			Social	& Virtual         &  Guo et al.~\cite{guo2023suspicion}                 &    Economic system:  game theory               \\ \hline
			Social	& Virtual       &  Zhao et al.~\cite{zhao2023competeai}                 &    Economic system: consumption market                     \\ \hline
			Social	& Virtual            &  Han et al.~\cite{han2023guinea}                 &    Economic system: consumption market               \\ \hline
			Social	& Virtual        &  Zhao et al.~\cite{nascimento2023self}                 &   Economic system: consumption market                    \\ \hline
			Social	& Virtual           &  Chen et al.~\cite{chen2023put}                 &    Economic system:   auction market                   \\ \hline
			% Physical & Real   &    Gurnee et al.~\cite{gurnee2023language}               &    Perception of space and time    \\ \hline
			Physical & Real         & Shah et al.~\cite{shah2023lm}      &    Navigation behavior  \\ \hline
			% Physical & Real        &   LLM-Planner~\cite{song2023llm}    &    Planning    \\ \hline
			Physical & Real      &   NLMap~\cite{chen2023open}    &    Navigation behavior   \\ \hline
			Physical & Real         &   Zou et al.~\cite{zou2023wireless}   &   Wireless network users \\ \hline
			Physical & Real        &   Cui et al.~\cite{cui2023drive}  & Vehicle drivers   \\ \hline
			Physical & Virtual & GITM~\cite{zhu2023ghost} 		& Tool-usage simulation in sandbox game  \\ \hline	
			Cyber	& Real     & WebAgent~\cite{gur2023real} & Human behaviors in Web \\ \hline
			Cyber	& Real     & Mind2Web~\cite{deng2023mind2web} & Human behaviors in Web \\  \hline
			Cyber	& Real     & Zhou~\textit{et al.}~\cite{zhou2023webarena}& Human behaviors in Web \\  \hline
			Cyber	& Real     & Park~\textit{et al.}~\cite{park2023choicemates}& Human behaviors in Web \\  \hline
			Cyber	& Virtual           & RecAgent~\cite{wang2023recagent} & Interaction with recommender system      \\ \hline
			Cyber	& Virtual           & Agent4Rec~\cite{zhang2023generative} & Interaction with recommender system            \\ \hline
			Hybrid	& Virtual          & Williams et al.~\cite{williams2023epidemic} & Epidemic spreading     \\ \hline
			Hybrid	& Virtual                  & Generative Agents~\cite{park2023generative} & Sandbox social life \\ \hline
			Hybrid & Real & WarAgent~\cite{hua2023war} 		& War simulation \\ \hline	
			Hybrid & Real           &  Li et al.~\cite{li2023large}                 &    Economic system: macroeconomics                 \\ \hline
			Hybrid	& Real     & UGI~\cite{ugi} & Human behaviors in real-world city \\ \hline
			
		\end{tabular}
	\end{table}
	
	\subsection{Social domain \uppercase\expandafter{\romannumeral1}: social sciences}

This section discusses the application of LLM agent-based modeling and simulation in social sciences.
Specifically, the existing works examine and explore LLM agents' effectiveness in replicating human behaviors and interactions and their role in validating social theories. They focus on how LLM agents can serve as tools for understanding complex social dynamics, enhancing collaborative problem-solving, etc., offering insights into both individual and collective social behaviors.

\subsubsection{Simulation of social network dynamics}

The part discusses whether LLM Agents, due to their human-like behavior, can be used to recreate and validate established social laws and patterns. This involves an analysis of how closely these agents can mimic human behavior and whether their actions can be quantified to validate or challenge existing theories in social science.

$S^3$~\cite{gao2023s} utilizes LLM-empowered agents to simulate individual and collective behaviors within a social network. This system effectively replicates human behaviors, including emotion, attitude, and interaction behaviors. It leverages real-world social network data to initialize the simulation environment, where information influences users' emotions and subsequent behaviors. The study particularly focuses on scenarios of gender discrimination and nuclear energy, demonstrating the ability of LLMs to simulate complex social dynamics. The results underscore LLM's ability to capture real-world social phenomena. Williams~\textit{et al.}~\cite{williams2023epidemic} study whether LLM agents can accurately reproduce the trend of epidemic spread. The results show that the LLM agent-based simulation system can replicate complex phenomena observed in the real world, such as multi-peak patterns. 

With \textit{Werewolf Game} as the environment, Xu~\textit{et al.}~\cite{xu2023exploring} examine LLMs' capabilities in simulating individual and collective behaviors in such a rule-based social environment. It reveals that LLMs can effectively engage in strategic social interactions, generating behaviors such as trust and confrontation, thus offering insights into their potential for social simulations. Acerbi~\textit{et al.}~\cite{acerbi2023large} demonstrate that the information transmitted by large language models mirrors the biases inherent in human social communication. Specifically, LLM exhibits preferences for stereotype-consistent, negative, socially oriented, and threat-related content, reflecting biases in its training data. The observation underscores that LLMs are not neutral agents; instead, they echo and potentially amplify existing human biases, shaping the information they generate and transmit. 
Zhang \textit{et al.}~\cite{zhang2023exploring} studies the impact of collaboration strategies on the performance of LLM agents. Specifically, three agents with distinct personalities (easy-going or overconfident) formed four different societies, employing eight collaboration strategies over three rounds to solve mathematical problems. It means that strategies initiating a debate show better results than those relying solely on memory reflection. That is, it highlights LLM agents' capability to exhibit human-like social phenomena of conformity and the Wisdom of Crowds effect, where collective intelligence tends to surpass individual capabilities. Kim \textit{et al.}~\cite{kim2023ai} assesses the boundaries and effectiveness of LLMs in modeling personal actions and societal dynamics, shedding light on their applicability for believable social simulations. Suzuki \textit{et al.}~\cite{suzuki2023evolutionary} and Zarza \textit{et al.}~\cite{de2023emergent} 
construct simulation systems with multiple agents, employing LLM as a generator of social strategy variations to simulate changes in cooperation/selfish strategies among agents in social cooperation and variations in social network structures, among other factors. 
Park \textit{et al.}~\cite{park2022social} investigates LLM agents' capacity to simulate online behaviors within forums. It demonstrates how LLMs can predict user interactions and responses by generating scenarios based on specific forum rules and descriptions. This simulation assists in refining forum regulations, highlighting the potential of LLMs in understanding and shaping digital social environments.

\begin{figure}[h]
	\centering
	\includegraphics[width=0.80\textwidth]{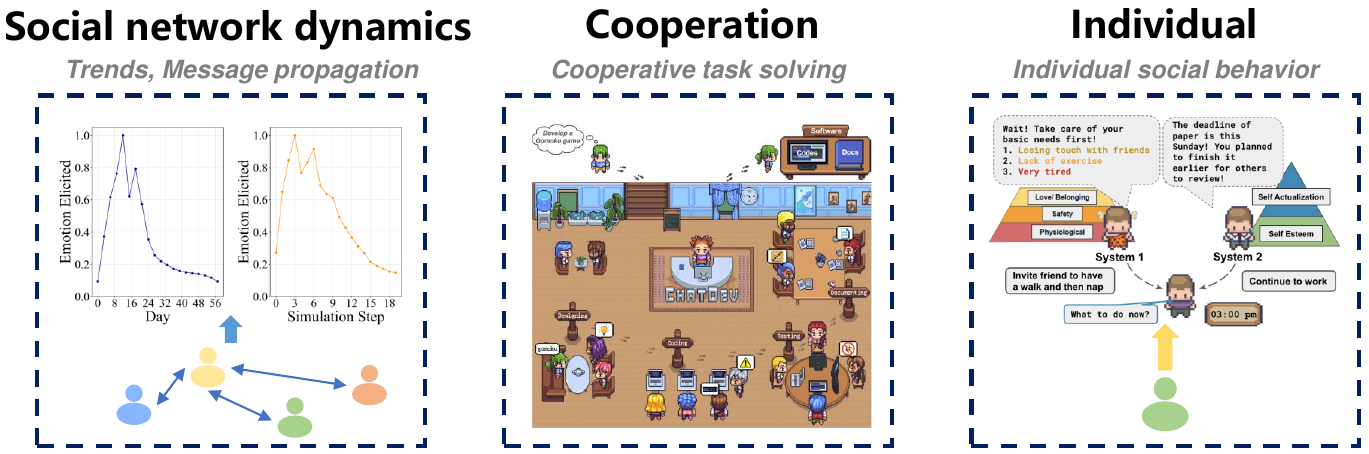}
	\caption{Taxonomy of LLM-based modeling and simulation in social sciences. Some materials are from S3~\cite{gao2023s}, ChatDev~\cite{qian2023communicative}, Humanoid Agents~\cite{wang2023humanoid}.}
	\label{fig:economy}
\end{figure}

\subsubsection{Simulation of cooperation}

Some other works pay attention to the human collaboration replicated by LLM agents. Specifically, they focus on how these agents, assigned distinct roles and functions, can mimic the cooperative behaviors observed in real human societies. The mechanisms and cooperative frameworks designed for these agents can enable them to work together efficiently toward achieving goals.

COLA~\cite{lan2023stance} proposed to organize LLM agents to discuss and finally decide the stance of social media text, with three role-played agents: analyzer, debater, and summarizer. The analyzers dissect texts from linguistic, domain-specific, and social media perspectives; the debaters propose logical links between text features and stances; finally, the summarizer considers all these discussions and determines the text's stance. The framework achieves SOTA performance on stance detection tasks. 
MAD~\cite{liang2023encouraging} proposed to use LLM agents to engage in reasoning-intensive question answering through structured debates. LLMs adopt the roles of opposing debaters, each arguing for a different perspective on the solution's correctness. MAD enforces a ``tit for tat'' debate dynamic, wherein each agent must argue against the other's viewpoint, leading to a more comprehensive exploration of potential solutions. A judge agent then evaluates these arguments to arrive at a final conclusion. This work fosters divergent thinking and deep contemplation, addressing the degeneration-of-thought issue common in self-reflection methods. 
CHATDEV~\cite{qian2023communicative}
is a virtual software development company where LLM Agents collaborate to develop computer software, with different roles for agents including CEO, CTO, designers, and programmers. The cooperation process encompasses designing, coding, testing, and documenting, with agents engaging in role-specific tasks like brainstorming, code development, GUI designing, and documentation.
% The framework achieves promising performance on software producing. 
MetaGPT~\cite{hong2023metagpt} also introduces a novel framework for collaborative software development with LLM agents, simulating a software company. 
The roles the agents play including Product Manager, Architect, Project Manager, Engineer, and QA Engineer, which follows the standardized operating procedures. 
Each role contributes sequentially, from requirement analysis and system design to task distribution, coding, and quality assurance, showcasing LLMs' potential in efficiently mimicking human cooperative behaviors and workflows in complex software development. 
ChatEval~\cite{chan2023chateval} presents a multi-agent framework for text quality evaluation, employing LLMs as diverse role-playing agents, in which key roles include the public, critics, journalists, philosophers, and scientists, each contributing unique perspectives. 
The agents engage in sequential debates, with access to all communication history. Finally, a judge gives a final decision. It results in more accurate and human-aligned evaluations compared to single-agent methods. CAMEL~\cite{li2023camel} introduces a cooperative role-playing framework with communicative agents, focusing on tool development. It involves roles like Task Detailing Assistant, Commander, and Executor. Specifically, Task Detailing Assistant specifies tasks in detail, Commander provides step-by-step instructions based on these specifics, and Executor carries out these instructions.

The above efforts involve designing specific types of agents, their roles, and the collaboration framework for certain tasks. The limitation lies in their lack of versatility, as the design of the agents is not flexible or adaptable. 
To address it, some work focuses on adaptively performing tasks with automated generated LLM agents and cooperation framework. AgentVerse~\cite{chen2023agentverse} simulate human group problem-solving focusing on adaptively generating LLM agents for diverse tasks. It involves four stages: 1) Expert Recruitment, where agent composition is determined and adjusted; 2) Collaborative Decision-Making, where agents plan problem-solving strategies; 3) Action Execution, where agents implement these strategies; 4) Evaluation, assessing progress, and guiding improvements. 
That is, it can effectively enhance agents' capabilities across various tasks, from coding to embodied AI, demonstrating their versatility in collaborative problem-solving. 
Wang~\textit{et al.}~\cite{wang2023unleashing} introduce Solo Performance Prompting (SPP) to emulate human-like cognitive synergy, which transforms a single LLM into a multi-persona agent, enhancing problem-solving in tasks requiring complex reasoning and domain knowledge.
For tasks like trivia creative writing and logic grid puzzles, SPP significantly outperforms standard methods, showcasing its effectiveness in collaborative problem-solving.
CoELA~\cite{zhang2023building} integrates LLMs' critical capabilities, including natural language processing, reasoning, and communication, into a novel cognitive-inspired modular framework.
The authors evaluate CoELA in various embodied environments like C-WAH and TDW-MAT, demonstrating its proficiency in perceiving, reasoning, communicating, and planning. The results show that CoELA surpasses traditional planning methods and exhibits effective cooperation and communication behaviors. 

In conclusion, simulating collaborative behaviors among LLM agents in various frameworks has shown their potential in emulating human cooperative behaviors to tackle a wide range of problem-solving tasks.

\subsubsection{Simulation of individual social behavior}
In the simulation of social dynamics and cooperative problem-solving, LLM agents show a strong ability to replicate human behavior. 
However, achieving a closer approximation to real human responses from the individual perspective is also of great significance.
In this section, we discuss how the recent works approach the problem how to better simulate the individual human behavior in a social context with LLM agents, enhancing their decision-making processes, interaction patterns, and emotional responses. 

Humanoid Agents~\cite{wang2023humanoid} propose a novel approach to enhancing the realism of LLM agent simulations. By incorporating elements of human cognitive processing (System 1~\cite{daniel2017thinking}), such as basic needs, emotions, and relational closeness, Humanoid Agents are designed to behave more like humans. These dynamic elements allow agents to adapt their activities and interactions based on their internal states, thereby bridging the gap between simulated and real human behavior. The platform also facilitates immersive visualization and analysis of these behaviors, advancing the field of social simulation and cooperative problem-solving. This approach demonstrates a significant leap in individual agent design, moving closer to replicating the complexities of human decision-making and interaction patterns.
SocioDojo~\cite{liu2023sociodojo} is a lifelong learning environment using real-world data for training agents in societal analysis and decision-making. It introduces an innovative Analyst-Assistant-Actuator framework and Hypothesis-Proof prompting, resulting in notable improvements on the time series forecasting task.
Liu~\textit{et al.}~\cite{liu2023training} presents a novel approach to optimizing LLM agents by refining agents' decision-making processes, interaction patterns, and emotional responses through a three-stage alignment learning framework, Stable Alignment. This framework, which efficiently teaches social alignment to LLMs, is based on simulated social interactions, detailed feedback, and progressive refinement of responses by autonomous social agents. 

Moreover, some studies use LLM agents to simulate human responses in social science research.
Argyle \textit{et al.}~\cite{argyle2023out} use LLM agents as proxies for specific human populations to generate responses in social science research. The authors show that, conditioned on socio-demographic profiles, LLM agents can generate outputs similar to human counterparts.
Hamalainen \textit{et al.}~\cite{hamalainen2023evaluating} construct LLM agents to simulate real participants to fill in open-ended questionnaires and analyze the similarity between the response and real data. The results show that synthetic responses generated by large language models cannot be easily distinguished from human data.
In short, these works indicate that LLM agents can be useful in social science experiments to simulate human responses with much lower costs. 

Some researchers have also studied LLM agents' ability to simulate human behavior in social psychological experiments.
Specifically, they~\cite{singh2023mind,binz2023using} use psychological tests to simulate the human response to test the cognitive ability, emotional intelligence~\cite{elyoseph2023chatgpt}, and psychological well-being~\cite{li2022gpt} of LLMs, demonstrating that LLM agents have human-like intelligence to a certain degree.

\subsection {Social Domain \uppercase\expandafter{\romannumeral2}: Economic System}
\begin{figure}[t!]
\centering
\includegraphics[width=0.80\textwidth]{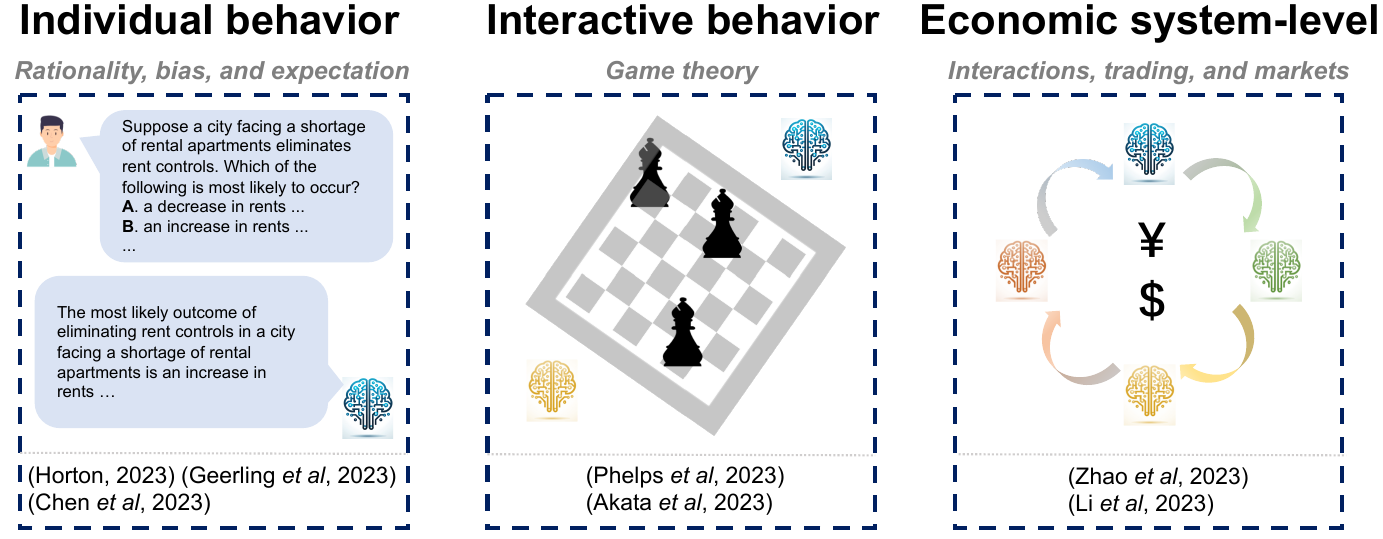}
\caption{Taxonomy of LLM-based modeling and simulation in economic systems.}
\label{fig:economy}
\end{figure}

This section discusses another important field in the social domain, the economic system. Currently, LLM-driven economic simulations can be categorized into three types based on the number of agents involved: individual behavior, interactive behavior, and economic system-level simulations. For individual behavior simulations, the primary goal of related research is to simulate the human-like economic decision-making capabilities of LLMs~\cite{horton2023large,geerling2023chatgpt,bauer2023decoding,chen2023emergence} or their understanding of economic phenomena~\cite{bybee2023surveying,xie2023wall,faria2023artificial}. This provides an empirical foundation for the latter two types of economic simulations and is currently a more extensively researched area. In interactive behavior simulations, the focus is mainly on game theory, exploring widely focused behaviors of LLMs during game-playing, such as cooperative and reasoning behaviors~\cite{phelps2023investigating,anonymous2023large,akata2023playing,guo2023suspicion}. For system-level simulations, the research primarily targets market simulations, such as consumption markets or auction markets, and investigates the rationality or optimality of LLMs' economic behaviors within these markets~\cite{zhao2023competeai,anonymous2023rethinking,chen2023put,li2023large}. The illustration is shown in Figure \ref{fig:economy}.

\subsubsection{Individual economic behavior simulation}
Considering the human-like characteristics of LLMs, many researchers attempted to replace humans in behavioral economics experiments with LLMs to observe the rational and irrational factors in their economic decision-making. Horton~\cite{horton2023large} replicated classic behavioral economics experiments using LLMs, including unilateral dictator games, fairness constraints~\cite{kahneman1986fairness}, and status quo bias~\cite{samuelson1988status}, confirming the human-like nature of LLMs in aspects such as altruism, fairness preferences, and status quo bias~\cite{horton2023large}. Although the experiment was conducted simply by asking GPT questions and analyzing responses, this represents a preliminary attempt to explore the use of LLMs for simulating human economic behavior. Chen \textit{et al.}~\cite{chen2023emergence} have employed standard frameworks, \textit{revealed preference theory}, to simulate the rationality in the economic decisions of GPTs. Results show that GPT performs largely rationally in risk, time, social, and food preferences domains in terms of budgetary decisions. Additionally, Geerling \textit{et al.}~\cite{geerling2023chatgpt} have utilized the Test of Understanding in College Economics to simulate LLMs' comprehension of microeconomics and macroeconomics, with results indicating that LLMs outperform most students who have taken economics courses.

Another research line to test the economic capabilities of LLMs involves accurately understanding certain socio-economic phenomena, specifically by using external text information (such as news) to predict future economic changes. Xie \textit{et al.}~\cite{xie2023wall} used LLM to predict stock market movements with historical stock data and related tweets based on the perception of investor sentiment. However, the predictive performance of LLMs is worse than that of state-of-the-art methods, and in some cases, it is even inferior to traditional linear regression. Faria \textit{et al.}~\cite{faria2023artificial} utilized LLMs for quarterly inflation forecasts, achieving accuracy comparable to, if not surpassing, the results of the Survey of Professional Forecasters~(SPF). Bybee \textit{et al.}~\cite{bybee2023surveying} tested LLMs on their predictions of finance and macroeconomics after reading specific sections of The Wall Street Journal, with results equivalent to those of SPF. These results suggest that LLMs possess a basic understanding of economic and financial markets but still lack sufficient and precise perception for accurate prediction, requiring more domain-specific data for additional fine-tuning.

\subsubsection{Interactive economic behavior simulation}
These simulations mainly focus on game theory, where there are only two or a few agents as opponents. Observing and analyzing the interactive behavior and capabilities of LLMs in various classic games is a current research hotspot.
Guo~\cite{guo2023gpt} studied the behavior of large language model agents in the ultimatum game and prisoner's dilemma game and found that the agents exhibit some similar patterns as humans, such as the positive correlation between offered amounts and acceptance rates in the ultimatum game.
Phelps \textit{et al.}~\cite{phelps2023investigating} found that incorporating individual preferences into prompts can influence the level of cooperation of LLMs. Specifically, they construct LLM agents with different personalities like competitive, altruistic, self-interested, etc., via prompts. Then, they let the agents play the repeated prisoner's dilemma game with bots with fixed strategies (e.g. always cooperate, always defect, or tit-for-tat) and analyze the agents' cooperation rate. They find that competitive and self-interested LLM agents show a lower cooperation rate, while altruistic agents demonstrate a higher cooperation rate, indicating the feasibility of constructing agents with different preferences through natural language. However, LLM agents also have limitations in some capabilities, such as the inability to reasonably respond to opponents' actions, which may lead to higher cooperation preferences with betraying opponents. 
The understanding of LLM social behaviors is very important for subsequent developments in artificial intelligence and its impact on human social behavior.
Other research~\cite{anonymous2023large} measured LLMs' rationality and strategic reasoning ability using the second-price auction and the Beauty Contest game. In such games, fully rational players are assumed to choose the most beneficial choice from their point of view, which results in the Nash equilibrium. Therefore, the authors define the deviation of LLMs' behavior from Nash equilibrium as the rationality degree. Moreover, they measure the strategic reasoning ability of LLMs by the ratio of the actual payoff to the optimal payoff. Experiments show that LLMs generally demonstrate rationality to some degree while they often cannot reach the Nash equilibrium. Among them, GPT-4 shows better strategic reasoning ability and can converge to Nash equilibrium faster than other LLMs like GPT-3.5 and text-davinci. The authors claim to provide a benchmark for testing the economic capabilities of the LLM research community.
Akata \textit{et al.}~\cite{akata2023playing} discovered through experiments in multiple game scenarios that LLMs are skilled in games valuing their self-interest but not as adept at coordinating with others. Specifically, in the prisoner's dilemma, GPT-4 will cooperate well with a cooperative opponent but will always choose to defect after the opponent defects once. In the Battle of the Sexes, GPT-4 cannot coordinate well with the opponent's choices to obtain maximum payoff.

In addition to the observations on the cooperative behavior and reasoning abilities of LLM agents during gaming, there are also a few studies attempting to construct strong game-playing agents. Guo \textit{et al.}~\cite{guo2023suspicion} goes beyond simple measurement and enhances LLMs' gaming abilities through prompt engineering. This work, specifically in an incomplete information game (namely Leduc Hold’em), has created agents with higher-order theory of mind that can significantly outperform traditional algorithm-based opponents without the requirement for training. Meta Fundamental AI Research Diplomacy Team (FAIR) \textit{et al.}~\cite{meta2022human} proposed the first AI agent Cicero combining a language model and reinforcement learning to play the Diplomacy game. After competing with real humans in online games anonymously, results show that Cicero can outperform 90\% of players. Even without employing LLMs, it has been demonstrated that earlier language models can approach or even surpass human capabilities in the realm of strategic gaming.

Moreover, Mao \textit{et al.}~\cite{mao2023alympics} developed a simulation framework named \textit{Alympics}, consisting of a sandbox playground and several agent players. The sandbox playground serves as the environment that stores and executes game settings, and agent players interact with the environment. The framework enables controlled, scalable, and reproducible simulation of game theory experiments.

The results from these simple simulation environments further validate the perception, reasoning, and planning capabilities of LLMs. In order to maximize their goals, LLMs consider their own benefits and opponents' strategies when making economic decisions. It is worth noting that these goals can be customized through prompts, such as maximizing returns or maximizing fairness.

\subsubsection{Economic system-level simulation}
In an economic system, agents often interact with each other, trade goods, and form a market. These agents may not be limited to individuals but can also represent entities such as companies and banks, as these are also important components of the market. Zhao \textit{et al.}~\cite{zhao2023competeai}, through simple consumption market simulations, uncovered competitive behaviors of LLM agents in managing restaurants, which are aligned with well-known sociological and economic theories. Specifically, the dish prices tend to be consistent with each other in the two simulated restaurants. Matthew effect also emerges during the simulation, \textit{i.e.}, one restaurant becomes more popular and popular while another has few consumers. Moreover, restaurants imitate competitors' behaviors, and at the same time, they try to make differentiation to attract more consumers. Similarly, Han \textit{et al.}~\cite{han2023guinea} studied the collusion between firms' price strategies. They simulated the product pricing process of two firms in a market environment~(\textit{i.e.}, Bertrand duopoly game) based on LLM. The results show that in the absence of communication, prices tend to approach the Bertrand equilibrium price. However, with communication, collusion between the companies tends to bring prices closer to the monopoly price. Nascimento \textit{et al.}~\cite{nascimento2023self} simulated a simple online book marketplace and observed interesting phenomena such as price negotiation between sellers and buyers. Some research~\cite{anonymous2023rethinking} attempts to have LLMs act as intermediaries in information trading markets to address the issue of information asymmetry between buyers and sellers. Specifically, when a seller presents information and quotes a price as the response to the query from a buyer, an LLM agent, acting as an intermediary, can decide whether to purchase and, if choosing not to, forget the information seen, thus protecting the seller's interests. In experiments, the information to exchange is actually the `passage' from documents on the topic of LLMs from ArXiv. The results show that LLM can not only make rational purchasing decisions in this information market but also ensure the rationality of the overall market dynamics; for example, a higher budget can improve the quality of purchased answers~(response to queries). Chen \textit{et al.}~\cite{chen2023put} have developed LLM agents with planning capabilities in constructed virtual auction markets to achieve higher profits given limited budgets. Experiments show that LLM has the crucial abilities to participate in the auction, including managing budgets, considering long-term returns, etc, even through only simple prompts.

\subsection {Physical domain}

For the physical domain, the applications for LLM agent-based modeling and simulation include mobility behaviors, transportation, wireless networks, etc.
\subsubsection{LLM agents for simulating mobility behaviors}

Understanding real-world space and time is crucial to harness LLMs for agent-based modeling and simulation in human mobility behaviors. Researchers have delved into this issue through various investigations~\cite{gurnee2023language,manvi2023geollm}. 
Gurnee~\textit{et al.}~\cite{gurnee2023language} focuses on probing LLMs to extract representations of real-world locations and temporal events, and the results demonstrate that these models build spatial and temporal representations in the neural layers. 
Manvi~\textit{et al.}~\cite{manvi2023geollm} delves into the geospatial knowledge embedded in LLMs. By fine-tuning LLMs on map-based prompts, substantial geospatial knowledge within LLMs is illustrated and shows improvements in tasks related to population density, asset wealth, and education. These investigations contribute valuable insights into the nuanced understanding of real-world space and time by LLMs, laying the groundwork for their application in agent-based simulations.

Based on their fundamental abilities, LLMs have showcased remarkable capabilities in simulation for the physical domain. For simulating the human-like navigation behaviors in the physical environment, LM-Nav~\cite{shah2023lm} combines large language models with image-language alignment algorithms. 
Following it, LLM-Planner~\cite{song2023llm} harnesses large language models to achieve few-shot planning for embodied agents. 
Moving into the domain of real-world planning with large language models, Chen~\textit{et al.}~\cite{chen2023open} introduces NLMap, which creates an open-vocabulary and queryable scene representation, allowing language models to gather and integrate contextual information for context-conditioned planning. Additionally, Shah~\textit{et al.}~\cite{shah2023gnm} study training a general goal-conditioned model to simulate human-like vision-based navigation, demonstrating the broad generalization capabilities of LLMs in complex physical environments.

\subsubsection{LLM agent-based modeling and simulation for transportation}
The possibility of using LLM agents for other applications in the physical domain, like transportation, has also been explored. 
Jin \textit{et al.}~\cite{jin2023surrealdriver} designs an LLM agent to simulate the driving behavior of human drivers. Specifically, the agent interacts with a simulator named CARLA, where it receives information about the state of the car and environment from the simulator and decide what to do next, such as stop, speed up, change lane, and so on, which will be fed back to the simulator. During the decision process, the agent will consider its recent behaviors using a memory module and also take into account safety criteria as well as guidelines learned from human expert drivers. Experiments show that the agent design can significantly reduce collision rate and make the agent's behavior more human-like. Moreover, the agent manages to perform complex driving tasks such as overtaking.

\subsubsection{LLM agent-based modeling and simulation for wireless network}
In addition, some researchers focus on deploying LLM agents to simulate device users in the city infrastructure, such as the wireless network. Zou \textit{et al.}~\cite{zou2023wireless} proposes a framework where multiple on-device LLM agents can interact with the environment and exchange knowledge to solve a complex task together.
Specifically, intents from humans or machines are provided to agents through wireless terminals, and the tasks are divided and planned collaboratively among multiple agents by leveraging the knowledge of different LLMs and device capabilities. On each device, the agent observes the environment and actors to execute decisions. On-device LLMs can extract semantic information from various data types and store it for future task planning. To deal with a specific task, the agent can retrieve relevant information or create lower-level tasks and send them to other agents to achieve the goal. The authors demonstrate the ability of the framework by an example of a wireless energy-saving task, where four users aim to reduce the network energy consumption while keeping the transmission rate. In the experiment, the agents gradually decrease their own power level based on previous actions of other users and manage to achieve the target after a few iterations, which shows the potential of LLM agent-based modeling and simulation in solving wireless network problems.

\subsection {Cyber domain}

	Agent-based modeling and simulation cyberspace mainly involves various human behaviors such as information access, website visitation, network attack/defense, etc., in cyberspace.
	
	WebAgent~\cite{gur2023real} is introduced as an LLM-driven agent capable of learning from its experiences to simulate human behaviors on real websites based on natural language instructions. It strategizes by breaking down instructions into manageable sub-parts, condenses lengthy HTML documents into relevant sections for the task at hand, and interacts with websites using Python programs derived from this information.
	Mind2Web~\cite{deng2023mind2web} further used large language models (LLMs) to construct these generalist web agents. While the sheer size of raw HTML from real websites poses a challenge for LLMs, Mind2Web demonstrates that pre-filtering this data with a smaller language model substantially enhances the effectiveness and efficiency of the LLMs in generating human-like web browsing behaviors.
	Zhou~\textit{et al.}~\cite{zhou2023webarena} further addressed the discrepancy between current language-guided autonomous agents, often tested in simplified synthetic environments, and the complexity of real-world scenarios.
	The authors build a highly realistic and reproducible environment specifically tailored for language-guided agents simulating human behaviors on the web.
	Park~\textit{et al.}~\cite{park2023choicemates} simulates the online decision-making scenarios, exploring the challenges individuals face when lacking domain expertise while searching for and making decisions using online information. 
	Wang~\textit{et al.}~\cite{wang2023recagent} proposed to build large language models to interact with recommender systems by selecting from recommendation results and providing positive or negative feedback. It serves as the testing protocol for evaluating the recommender system's performance: whether it can satisfy the agents' preferences well. 
	
	In RecAgent~\cite{wang2023recagent}, the researchers explore the potential of LLMs in simulating user behaviors within online environments, particularly recommender systems. By creating an LLM-based autonomous agent framework, the study investigates how these agents can simulate complex human interactions and decisions in a virtual environment. This approach enables a novel method for studying user behavior, offering insights into how users might react to different scenarios in digital platforms, thus advancing our understanding of user dynamics in virtual spaces.
	Zhang~\textit{et al.}~\cite{zhang2023generative} proposed to build generative agents for the recommender system in which the authors design LLM-empowered generative agents equipped with user profile, memory, and actions modules specifically tailored for the recommender system. The proposed agents can emulate the filter bubble effect and discover the underlying causal relationships in recommendation tasks.

	\subsection {Hybrid domain}
In some studies, simulations are conducted that simultaneously consider more than one domain, such as physical and social, and we refer to these simulations as being within a hybrid domain. 

As a pioneering work, Generative Agents~\cite{park2023generative} offers a compelling insight into the generation of believable individual and social behaviors. The research focuses on a central question: how can generative agents reliably produce human-like individual actions and social dynamics? It delves into an agent architecture that integrates memory (for storing past experiences), reflection (to make rational present decisions), and planning (for future actions). This architecture is critically evaluated through crowd-sourced assessments, affirming the effectiveness of the memory, reflection, and planning modules in generating rational behaviors. Notably, this approach led to complex social scenarios, such as Valentine's Day parties and mayoral elections, underscoring the agents' proficiency in simulating nuanced human interactions and societal events. This research offers a substantial contribution to social simulations, demonstrating the advanced potential of LLM agents in replicating the depth and complexity of human social behaviors. 

Williams~\textit{et al.}~\cite{williams2023epidemic} conduct an epidemic simulation within a hybrid domain. In this simulation, social relationships influenced individuals' perception of the epidemic, while individuals' physical movements within spatial contexts affected their susceptibility to infection. Welfare Diplomacy~\cite{mukobi2023welfare} sets a benchmark, a nation-to-nation war/welfare equilibrium tabletop game designed to evaluate the collaborative capabilities of large language models.

Hua~\textit{et al.}~\cite{hua2023war} proposed to use LLM agents to represent countries and simulate their decisions and consequences, based on which the historical international conflicts, including World War I, World War II, and the Warring States Period in Ancient China are selected for evaluation. In the LLM agent-based war simulations, the emergent interactions among countries help explain why the wars occur. 

Li \textit{et al.}~\cite{li2023large} simulate a hybrid macroeconomic system and expand the scale of simulation environments from tens to hundreds. Specifically, they simulate LLM-empowered agents' work and consumption behaviors in a macroeconomic market. The proposed perception, memory, and action modules endow the agents with real-world heterogeneity, the ability to grasp market dynamics, and decision-making considering multiple economic factors, respectively. Experimental results show the emergence of more reasonable and stable macroeconomic indicators~(price inflation, unemployment rate, GDP, and GDP growth rate) and regularities~(Phillips curve and Okun's law) compared with traditional rule-based ABM~\cite{lengnick2013agent,gatti2011macroeconomics} and RL-based approaches~\cite{zheng2022ai}. Especially, only the simulation based on LLM agents can produce the correct Phillips curve, \textit{i.e.,} negative relationship between the unemployment rate and inflation. This advantage is owned by LLM's accurate perception of market dynamics, such as the deflation of labor markets. 

Urban Generative Intelligence (UGI)~\cite{ugi} is a platform that constructs a real-world urban environment provided by digital twins, which provide various interfaces for embodied agents to generate many behaviors, supported by a foundation model named CityGPT, which is trained on city-specific multi-source data. In this platform, multiple categories of LLM-based agents can simulate human-like behaviors, including social interactions, economic activities, mobility, street navigation, etc., showing promising abilities in simulating city activities based on embodied agents.

\section{Open problems and future directions}

\subsection{Efficiency of Scaling Up}
Many studies of LLM agents find it advantageous to simultaneously simulate multiple personas and exploit the synergy effect by allowing them to communicate and vote for the final output~\cite{yao2023tree}. 
For example, researchers find LLM-based software development can be significantly improved by simulating a virtual software company with diverse social identities, including chief officers, professional programmers, test engineers, and art designers~\cite{qian2023communicative}. This virtual company is capable of streamlining the development of complex software solutions in the stages of designing, coding, testing, and documenting. Moreover, researchers generally find scaling up the number of simulated agents and deploying more diverse personas are beneficial in various tasks~\cite{zhuge2023mindstorms}.

However, simulating societies of large-scale LLM agents is very computationally expensive. Extensive research efforts are dedicated to optimizing the memory footprint~\cite{sheng2023high} and operation subroutines~\cite{aminabadi2022deepspeed} of language models.  
Researchers also develop several effective model compression techniques~\cite{zhu2023survey}, such as knowledge distillation and quantization.
In the context of LLM agent simulation, batch prompting~\cite{cheng2023batch} is a highly relevant technique that is capable of simulating multiple agents in batches. Experiments show batch prompting can achieve up to 5$\times$ efficiency improvement in inference token and time costs. Besides, MetaGPT is proposed to improve the efficiency of multi-agent collaboration in virtual software companies~\cite {hong2023metagpt}. They leverage a shared message pool and subscribe mechanism to reduce the time and token cost of generating one line of code. Despite the previous efforts of accelerating LLM agents, simulating large-scale LLM agents remains a highly challenging task, which significantly hinders LLM agent simulation from reaching its full potential. Simulating large societies of LLM agents not only can effectively improve the performance in downstream tasks, but also has the potential to mimic the emergence properties of human societies, and hence reveal the underlying mechanisms~\cite{caldarelli2023role}. Therefore, it is an important open problem to achieve full-process acceleration of LLM agent simulations. 

\subsection{Benchmark}

Benchmarks have significantly advanced the development of AI in the past decade. Landmark benchmarks like ImageNet~\cite{russakovsky2015imagenet}, GLUE~\cite{wang2018glue}, and the benchmarks in graph learning~\cite{hu2020open,dwivedi2023benchmarking} have been pivotal to the rapid innovation in the fields of computer vision, natural language processing and graph neural networks.

Recently, there has been a surge in benchmarks that assess the capabilities of LLM-driven agents, highlighting the growing interest in this emerging area. For example, researchers~\cite{valmeekam2022large} develop benchmarks to evaluate LLM's capability in planning and reasoning about change, focusing on symbolic models and structured inputs compatible with such representations. Meanwhile, AgentBench develops a multi-dimensional benchmark with 8 distinct environments to assess the capabilities of LLM-driven agents in various multi-turn open-ended generation settings~\cite {liu2023agentbench}. MLAgentBench, on the other hand, designs a suite of ML tasks for benchmarking LLM-driven AI research agents, including tasks like image classification and sentiment classification~\cite{huang2023benchmarking}. Researchers also propose to evaluate LLM-driven agents with embodied tasks, using them as high-level planners in robotics setups or in textual environments, focusing on the interaction between planning and action, like ALFWorld~\cite{shridhar2020alfworld} and ComplexWorld~\cite{basavatia2023complexworld}. 
On top of textual environments, online reinforcement learning approaches are developed to align LLM agents with human preference and evaluate their performance~\cite{carta2023grounding}.

However, the previous benchmarks mainly focus on the decision and planning capability of LLM-driven agents, the assessment of LLM-driven agent simulation is still inadequate. On the one hand, there still exist challenges in evaluating the performance of agent simulations. Previous works often examine the statistics feature of simulated behaviour~\cite{feng2020learning}, such as the spatial and temporal distribution. Recent studies also recruit human evaluators to gather feedback on the believability of the simulation~\cite{park2023generative}. However, developing benchmarks for quantitative and qualitative evaluation of LLM-driven agent-based simulation remains a largely open problem and a promising future research direction. On the other hand, LLM-driven simulation might serve as a realistic environment that provides high-quality feedback to train other AI models. For example, previous studies explore the simulations of social segregation~\cite{sert2020segregation}, competing firms~\cite{osoba2020policy}, competitive games~\cite{park2019multi}, and coordination of different stakeholders~\cite{bone2010simulation}. Such simulations can serve as a benchmark to train and evaluate the reinforcement learning models. A recent study~\cite{wu2023plan} proposes a PET framework to leverage LLM-driven agents as a supervisor of low-level trainable models, which simplifies challenging control tasks by translating task descriptions into high-level sub-tasks and then tracking the accomplishment of these sub-tasks. 
Additionally, more research efforts should be dedicated to the benchmarks of AI for social good~\cite{cowls2021definition}.

\subsection{Open Platform}

Building open platforms for LLM-driven agents will play a pivotal role in this emerging research area that could substantially reduce the barriers of LLM-driven ABS and foster a vibrant community, echoing the calls for open-source software~\cite{weber2004success} and open science~\cite{national2018open}. 
The recent advance of LLMs has led to the public releases of several powerful pretrained language models. For example, Bidirectional Encoder Representations from Transformers (BERT) has been publicly released and gained huge influences in the past few years~\cite{devlin2018bert}. GPT2, a predecessor to the current ChatGPT family, was released by OpenAI with limited model sizes for open-source use~\cite{radford2019language}.
Additionally, Meta AI recently released a collection of open foundation and fine-tuned chat models named LLaMa 2~\cite{touvron2023llama}, which range in scale from 7 billion to 70 billion parameters. 
These open-source LLMs demonstrate powerful capabilities in various natural language tasks, which can be further adapted for specific downstream tasks with efficient fine-tuning methods such as Low-Rank Adaptation (LoRA)~\cite{hu2021lora}.

The recent proliferation of LLM-driven agents has also resulted in several open-source platforms. Voyager is an example open-source framework of embodied LLM-driven agents, capable of continuously acquiring diverse skills and making novel discoveries in Minecraft without human intervention~\cite{xi2023rise}. 
Researchers also develop open-source frameworks for real-world task-solving agents, such as XAgent~\cite{xagent2023} that are designed as a general-purpose framework of automatic task-solving. Moreover, ModelScope-Agent~\cite{li2023modelscope} is proposed as a general and customizable agent framework designed for real-world applications, which supports model training on multiple open-source LLMs and offers diversified and comprehensive APIs. On top of the textual embodied environment ALFWorld, researchers developed BUTLER framework~\cite{shridhar2020alfworld} that can operate across text and embodied environments with three main components, \emph{i.e.}, brain, vision and body. This arrangement allows BUTLER to effectively bridge the gap between abstract language understanding and practical, embodied task execution in simulated virtual environments. However, these previous works mainly focus on task-solving LLM agents, while the open platforms for LLM-driven ABS are still lacking. Such gap can be largely attributed to the challenges of integrating LLM-driven agents with the complex environment of simulation. Urban Generative Intelligence (UGI)~\cite{ugi} is a recently proposed open platform that integrates embodied agents with the digital twins of cities, offering the opportunity to evaluate urban problems with large-scale urban agent simulations and solve them with multidisciplinary approaches. Despite this early attempt at urban system simulation, the development of an open platform for LLM-driven ABS is an emerging area that calls for more research attention.

\subsection{Robustness of LLM-driven ABS}
The robustness problems of LLM agent simulation can be classified into two main scenarios, adversarial attack and out-of-distribution generalization, which fundamentally stem from the robustness issues of the underlying language models~\cite{wang2023robustness}.
The current methodologies to address out-of-distribution generalization problems primarily resort to classic machine learning techniques~\cite{shen2021towards}, such as unsupervised representation learning, supervised model learning, and optimization methods.
As for the adversarial attack, various defense techniques have been proposed in recent studies. 
For example, researchers propose to certify LLM safety with an erase-and-check filter that detects adversarial prompts~\cite{kumar2023certifying}. 
Besides, moving target defense~\cite{chen2023jailbreaker} aims to select safe answers from the responses generated by different LLMs to enhance LLM system's robustness against jailbreaking attacks. Moreover, extensive benchmarks of adversarial prompts are formulated to evaluate LLM~\cite{zhu2023promptbench}.

As for the LLM agents, they often have tool-use capability~\cite{qin2023tool} and engage in human interactive scenarios, such as the conflict simulation actor that helps users learn conflict resolve through rehearsal~\cite{shaikh2023rehearsal}, which make the robustness of LLM agents have far-reaching consequences. Furthermore, in the context of multi-agent simulation, adversarial attacks might propagate among agents~\cite{tian2023evil}. More importantly, recent works show the simulations of multiple LLM agents show human-like collective behaviors~\cite{aher2023using,zhou2023sotopia}, such as social conformity and homophily, which could be exploited by adversaries as weaknesses in the societies of LLM agents. Improving the robustness of LLM agent simulation at both the individual and collective levels is an open problem.

\subsection{Ethical Risks in LLM Agents}
The advances of LLM unleash the unprecedented capability of human-like text generation and reasoning, raising concerns of potential ethical risks of misuse, such as jailbreaking~\cite {zhuo2023exploring}. For example, recent studies highlight the risks of generating malicious network payloads that could jeopardize cyber security at scale~\cite{charan2023text}, and emphasize the concerns of accuracy, recency, coherence, and transparency of LLM agents in medical practice~\cite{thirunavukarasu2023large}.
To gauge LLM agents' susceptibility to social bias and stereotype, researchers use semantic illusions and cognitive reflection tests~\cite{hagendorff2023human}, typically administered to human subjects, to quantify LLM's tendency to produce intuitive yet erroneous responses. They find early models from the GPT family have an increasing tendency to generate intuitive errors as their size increases, while ChatGPT-3.5 and 4 have a pattern shift that radically eliminates these errors and achieves superhuman accuracy. They speculate the pattern shift is driven by the employment of reinforcement learning from human feedback, a sophisticated technique only deployed in ChatGPT-3.5 and later models. These findings highlight the importance of embedding human preferences into the language models, instead of solely relying on web corpus.
In the context of LLM-driven agent simulations, researchers find when certain personas are assigned to ChatGPT it will generate output with 6$\times$ toxicity, engaging in discriminatory stereotypes,
harmful conversation, and offensive language~\cite{deshpande2023toxicity}.
Besides, a recent work~\cite{acerbi2023large} shows LLM agent exhibits human-like biases that prefer gender-stereotype-consistent, negative, and biologically counter-intuitive content. More importantly, such biases could be further amplified in the transmission chain in multi-agent settings.
The experimental results from previous studies emphasize the importance of ethical considerations in LLM-driven agent-based simulations, especially against the backdrop of the rapid proliferation of LLM agents in various domains.

Extensive efforts have been made to mitigate the potential ethical risks of LLM agents. A primary focus is to fundamentally align language models with human values~\cite{yi2023unpacking, yao2023instructions}. A recent survey classifies the alignment goals into three distinct levels, \emph{i.e.}, human instructions, human preferences, and human values. 
Besides, Moral Foundation theory is invoked to benchmark mainstream language models' alignment with the foundational ethical values of care, fairness, loyalty, authority, and sanctity~\cite{yi2023unpacking}.
Researchers also find LLM agents are susceptible to flattened caricatures when specific personas are assigned to them~\cite{cheng2023compost}. 
The CoMPosT framework is proposed to evaluate the multidimensionality of simulated LLM agents and provide a measure for caricature in LLM agent simulations. 
They find even the agents driven by the latest GPT-4 in the simulation of political and marginalized demographic groups.
Finally, to fundamentally address the potential ethical risks, many scholars advocate enhancing the interpretability of LLM agents, questioning the falsifiability of any moral principles learned by black box LLM agents~\cite{vijayaraghavan2023minimum}. Therefore, they propose to benchmark and continuously improve LLM agents' interpretability~\cite{zhao2023explainability}.

\section{Conclusion}\label{sec::conclusion}
Agent-based modeling and simulation is one of the most important methods to model complex systems in various domains.
The recent advances in large language models have reshaped the paradigm of agent-based modeling and simulation, providing a new perspective for constructing intelligent human-like agents rather than those driven by simple rules or limited-intelligence neural models.
In this paper, we take the first step to provide a survey of the agent-based modeling and simulation with large language models.
We systematically analyze why the LLM agents are required for agent-based modeling and simulation and how to address the critical challenges.
Afterward, we extensively summarize the existing works in four domains: cyber, physical, social, and hybrid, carefully describing how to design the simulation environment, how to construct the large language model-empowered agents, and what to observe and achieve based on agent-based simulation.
Lastly, given the unresolved limitations of existing works and this new and fast-growing area, we discuss the open problems and point out the important research directions, which we hope can inspire future research.
\bibliographystyle{plain}

%\bibliography{bibliography}

\begin{thebibliography}{100}
	
	\bibitem{acerbi2023large}
	Alberto Acerbi and Joseph~M Stubbersfield.
	\newblock Large language models show human-like content biases in transmission
	chain experiments.
	\newblock {\em Proceedings of the National Academy of Sciences},
	120(44):e2313790120, 2023.
	
	\bibitem{aher2023using}
	Gati~V Aher, Rosa~I Arriaga, and Adam~Tauman Kalai.
	\newblock Using large language models to simulate multiple humans and replicate
	human subject studies.
	\newblock In {\em International Conference on Machine Learning}, pages
	337--371. PMLR, 2023.
	
	\bibitem{akata2023playing}
	Elif Akata, Lion Schulz, Julian Coda-Forno, Seong~Joon Oh, Matthias Bethge, and
	Eric Schulz.
	\newblock Playing repeated games with large language models.
	\newblock {\em arXiv preprint arXiv:2305.16867}, 2023.
	
	\bibitem{alluhaybi2019survey}
	Bandar Alluhaybi, Mohamad~Shady Alrahhal, Ahmed Alzhrani, and Vijey
	Thayananthan.
	\newblock A survey: agent-based software technology under the eyes of cyber
	security, security controls, attacks and challenges.
	\newblock {\em International Journal of Advanced Computer Science and
		Applications (IJACSA)}, 10(8), 2019.
	
	\bibitem{aminabadi2022deepspeed}
	Reza~Yazdani Aminabadi, Samyam Rajbhandari, Ammar~Ahmad Awan, Cheng Li, Du~Li,
	Elton Zheng, Olatunji Ruwase, Shaden Smith, Minjia Zhang, Jeff Rasley, et~al.
	\newblock Deepspeed-inference: enabling efficient inference of transformer
	models at unprecedented scale.
	\newblock In {\em SC22: International Conference for High Performance
		Computing, Networking, Storage and Analysis}, pages 1--15. IEEE, 2022.
	
	\bibitem{an2012modeling}
	Li~An.
	\newblock Modeling human decisions in coupled human and natural systems: Review
	of agent-based models.
	\newblock {\em Ecological modelling}, 229:25--36, 2012.
	
	\bibitem{anonymous2023large}
	Anonymous.
	\newblock Large language models as rational players in competitive economics
	games.
	\newblock In {\em Submitted to The Twelfth International Conference on Learning
		Representations}, 2023.
	\newblock under review.
	
	\bibitem{anonymous2023rethinking}
	Anonymous.
	\newblock Rethinking the buyer{\textquoteright}s inspection paradox in
	information markets with language agents.
	\newblock In {\em Submitted to The Twelfth International Conference on Learning
		Representations}, 2023.
	\newblock under review.
	
	\bibitem{antonini2006discrete}
	Gianluca Antonini, Michel Bierlaire, and Mats Weber.
	\newblock Discrete choice models of pedestrian walking behavior.
	\newblock {\em Transportation Research Part B: Methodological}, 40(8):667--687,
	2006.
	
	\bibitem{argyle2023out}
	Lisa~P Argyle, Ethan~C Busby, Nancy Fulda, Joshua~R Gubler, Christopher
	Rytting, and David Wingate.
	\newblock Out of one, many: Using language models to simulate human samples.
	\newblock {\em Political Analysis}, 31(3):337--351, 2023.
	
	\bibitem{arora2023have}
	Daman Arora, Himanshu~Gaurav Singh, et~al.
	\newblock Have llms advanced enough? a challenging problem solving benchmark
	for large language models.
	\newblock {\em arXiv preprint arXiv:2305.15074}, 2023.
	
	\bibitem{arsanjani2013spatiotemporal}
	Jamal~Jokar Arsanjani, Marco Helbich, and Eric de~Noronha~Vaz.
	\newblock Spatiotemporal simulation of urban growth patterns using agent-based
	modeling: The case of tehran.
	\newblock {\em Cities}, 32:33--42, 2013.
	
	\bibitem{arthur1991designing}
	W~Brian Arthur.
	\newblock Designing economic agents that act like human agents: A behavioral
	approach to bounded rationality.
	\newblock {\em The American economic review}, 81(2):353--359, 1991.
	
	\bibitem{liu2023sociodojo}
	Anonymous Authors.
	\newblock Sociodojo: Building lifelong analytical agents.
	
	\bibitem{banisch2012agent}
	Sven Banisch, Ricardo Lima, and Tanya Ara{\'u}jo.
	\newblock Agent based models and opinion dynamics as markov chains.
	\newblock {\em Social Networks}, 34(4):549--561, 2012.
	
	\bibitem{barbosa2011simulation}
	Jos{\'e} Barbosa and Paulo Leit{\~a}o.
	\newblock Simulation of multi-agent manufacturing systems using agent-based
	modelling platforms.
	\newblock In {\em 2011 9th IEEE International Conference on Industrial
		Informatics}, pages 477--482. IEEE, 2011.
	
	\bibitem{barnes2013applications}
	Sean Barnes, Bruce Golden, and Stuart Price.
	\newblock Applications of agent-based modeling and simulation to healthcare
	operations management.
	\newblock In {\em Handbook of healthcare operations management: methods and
		applications}, pages 45--74. Springer, 2013.
	
	\bibitem{barros2004urban}
	Joana~Xavier Barros.
	\newblock {\em Urban Growth in Latin American Cities-Exploring urban dynamics
		through agent-based simulation}.
	\newblock University of London, University College London (United Kingdom),
	2004.
	
	\bibitem{basavatia2023complexworld}
	Shreyas Basavatia, Shivam Ratnakar, and Keerthiram Murugesan.
	\newblock Complexworld: A large language model-based interactive fiction
	learning environment for text-based reinforcement learning agents.
	\newblock In {\em International Joint Conference on Artificial Intelligence},
	2023.
	
	\bibitem{batty2003agent}
	Michael Batty.
	\newblock Agent-based pedestrian modelling.
	\newblock 2003.
	
	\bibitem{bauer2023decoding}
	Kevin Bauer, Lena Liebich, Oliver Hinz, and Michael Kosfeld.
	\newblock Decoding gpt’s hidden ‘rationality’of cooperation.
	\newblock 2023.
	
	\bibitem{beheshti2017comparing}
	Rahmatollah Beheshti, Mehdi Jalalpour, and Thomas~A Glass.
	\newblock Comparing methods of targeting obesity interventions in populations:
	an agent-based simulation.
	\newblock {\em SSM-population health}, 3:211--218, 2017.
	
	\bibitem{bellegarda2004statistical}
	Jerome~R Bellegarda.
	\newblock Statistical language model adaptation: review and perspectives.
	\newblock {\em Speech communication}, 42(1):93--108, 2004.
	
	\bibitem{beltran2017agent}
	Roxanne~S Beltran, J~Ward Testa, and Jennifer~M Burns.
	\newblock An agent-based bioenergetics model for predicting impacts of
	environmental change on a top marine predator, the weddell seal.
	\newblock {\em Ecological Modelling}, 351:36--50, 2017.
	
	\bibitem{binz2023using}
	Marcel Binz and Eric Schulz.
	\newblock Using cognitive psychology to understand gpt-3.
	\newblock {\em Proceedings of the National Academy of Sciences},
	120(6):e2218523120, 2023.
	
	\bibitem{bohlmann2010effects}
	Jonathan~D Bohlmann, Roger~J Calantone, and Meng Zhao.
	\newblock The effects of market network heterogeneity on innovation diffusion:
	An agent-based modeling approach.
	\newblock {\em Journal of Product Innovation Management}, 27(5):741--760, 2010.
	
	\bibitem{boiko2023emergent}
	Daniil~A Boiko, Robert MacKnight, and Gabe Gomes.
	\newblock Emergent autonomous scientific research capabilities of large
	language models.
	\newblock {\em arXiv preprint arXiv:2304.05332}, 2023.
	
	\bibitem{bone2010simulation}
	Christopher Bone and Suzana Dragi{\'c}evi{\'c}.
	\newblock Simulation and validation of a reinforcement learning agent-based
	model for multi-stakeholder forest management.
	\newblock {\em Computers, Environment and Urban Systems}, 34(2):162--174, 2010.
	
	\bibitem{bran2023chemcrow}
	Andres~M Bran, Sam Cox, Andrew~D White, and Philippe Schwaller.
	\newblock Chemcrow: Augmenting large-language models with chemistry tools.
	\newblock {\em arXiv preprint arXiv:2304.05376}, 2023.
	
	\bibitem{brown2006effects}
	Daniel~G Brown and Derek~T Robinson.
	\newblock Effects of heterogeneity in residential preferences on an agent-based
	model of urban sprawl.
	\newblock {\em Ecology and society}, 11(1), 2006.
	
	\bibitem{brown2020language}
	Tom Brown, Benjamin Mann, Nick Ryder, Melanie Subbiah, Jared~D Kaplan, Prafulla
	Dhariwal, Arvind Neelakantan, Pranav Shyam, Girish Sastry, Amanda Askell,
	et~al.
	\newblock Language models are few-shot learners.
	\newblock {\em Advances in neural information processing systems},
	33:1877--1901, 2020.
	
	\bibitem{bybee2023surveying}
	Leland Bybee.
	\newblock Surveying generative ai's economic expectations.
	\newblock {\em arXiv preprint arXiv:2305.02823}, 2023.
	
	\bibitem{cabrera2011optimization}
	Eduardo Cabrera, Manel Taboada, Ma~Luisa Iglesias, Francisco Epelde, and Emilio
	Luque.
	\newblock Optimization of healthcare emergency departments by agent-based
	simulation.
	\newblock {\em Procedia computer science}, 4:1880--1889, 2011.
	
	\bibitem{caldarelli2023role}
	G~Caldarelli, E~Arcaute, M~Barthelemy, M~Batty, C~Gershenson, D~Helbing,
	S~Mancuso, Y~Moreno, JJ~Ramasco, C~Rozenblat, et~al.
	\newblock The role of complexity for digital twins of cities.
	\newblock {\em Nature Computational Science}, pages 1--8, 2023.
	
	\bibitem{carta2023grounding}
	Thomas Carta, Cl{\'e}ment Romac, Thomas Wolf, Sylvain Lamprier, Olivier Sigaud,
	and Pierre-Yves Oudeyer.
	\newblock Grounding large language models in interactive environments with
	online reinforcement learning.
	\newblock {\em arXiv preprint arXiv:2302.02662}, 2023.
	
	\bibitem{chan2023chateval}
	Chi-Min Chan, Weize Chen, Yusheng Su, Jianxuan Yu, Wei Xue, Shanghang Zhang,
	Jie Fu, and Zhiyuan Liu.
	\newblock Chateval: Towards better llm-based evaluators through multi-agent
	debate.
	\newblock {\em arXiv preprint arXiv:2308.07201}, 2023.
	
	\bibitem{chang2023survey}
	Yupeng Chang, Xu~Wang, Jindong Wang, Yuan Wu, Kaijie Zhu, Hao Chen, Linyi Yang,
	Xiaoyuan Yi, Cunxiang Wang, Yidong Wang, et~al.
	\newblock A survey on evaluation of large language models.
	\newblock {\em arXiv preprint arXiv:2307.03109}, 2023.
	
	\bibitem{charan2023text}
	PV~Charan, Hrushikesh Chunduri, P~Mohan Anand, and Sandeep~K Shukla.
	\newblock From text to mitre techniques: Exploring the malicious use of large
	language models for generating cyber attack payloads.
	\newblock {\em arXiv preprint arXiv:2305.15336}, 2023.
	
	\bibitem{chen2023jailbreaker}
	Bocheng Chen, Advait Paliwal, and Qiben Yan.
	\newblock Jailbreaker in jail: Moving target defense for large language models.
	\newblock {\em arXiv preprint arXiv:2310.02417}, 2023.
	
	\bibitem{chen2023open}
	Boyuan Chen, Fei Xia, Brian Ichter, Kanishka Rao, Keerthana Gopalakrishnan,
	Michael~S Ryoo, Austin Stone, and Daniel Kappler.
	\newblock Open-vocabulary queryable scene representations for real world
	planning.
	\newblock In {\em 2023 IEEE International Conference on Robotics and Automation
		(ICRA)}, pages 11509--11522. IEEE, 2023.
	
	\bibitem{chen2023put}
	Jiangjie Chen, Siyu Yuan, Rong Ye, Bodhisattwa~Prasad Majumder, and Kyle
	Richardson.
	\newblock Put your money where your mouth is: Evaluating strategic planning and
	execution of llm agents in an auction arena.
	\newblock {\em arXiv preprint arXiv:2310.05746}, 2023.
	
	\bibitem{chen2012agent}
	Liang Chen.
	\newblock Agent-based modeling in urban and architectural research: A brief
	literature review.
	\newblock {\em Frontiers of Architectural Research}, 1(2):166--177, 2012.
	
	\bibitem{chen2023agentverse}
	Weize Chen, Yusheng Su, Jingwei Zuo, Cheng Yang, Chenfei Yuan, Chen Qian,
	Chi-Min Chan, Yujia Qin, Yaxi Lu, Ruobing Xie, et~al.
	\newblock Agentverse: Facilitating multi-agent collaboration and exploring
	emergent behaviors in agents.
	\newblock {\em arXiv preprint arXiv:2308.10848}, 2023.
	
	\bibitem{chen2023emergence}
	Yiting Chen, Tracy~Xiao Liu, You Shan, and Songfa Zhong.
	\newblock The emergence of economic rationality of gpt.
	\newblock {\em arXiv preprint arXiv:2305.12763}, 2023.
	
	\bibitem{cheng2023compost}
	Myra Cheng, Tiziano Piccardi, and Diyi Yang.
	\newblock Compost: Characterizing and evaluating caricature in llm simulations.
	\newblock {\em arXiv preprint arXiv:2310.11501}, 2023.
	
	\bibitem{cheng2023batch}
	Zhoujun Cheng, Jungo Kasai, and Tao Yu.
	\newblock Batch prompting: Efficient inference with large language model apis.
	\newblock {\em arXiv preprint arXiv:2301.08721}, 2023.
	
	\bibitem{chowdhery2023palm}
	Aakanksha Chowdhery, Sharan Narang, Jacob Devlin, Maarten Bosma, Gaurav Mishra,
	Adam Roberts, Paul Barham, Hyung~Won Chung, Charles Sutton, Sebastian
	Gehrmann, et~al.
	\newblock Palm: Scaling language modeling with pathways.
	\newblock {\em Journal of Machine Learning Research}, 24(240):1--113, 2023.
	
	\bibitem{cipi2011simulation}
	Eva Cipi and Betim Cico.
	\newblock Simulation of an agent based system behavior in a dynamic and
	unpredicted environment.
	\newblock {\em Simulation}, 1(4):172--176, 2011.
	
	\bibitem{conte2014agent}
	Rosaria Conte and Mario Paolucci.
	\newblock On agent-based modeling and computational social science.
	\newblock {\em Frontiers in psychology}, 5:668, 2014.
	
	\bibitem{cowls2021definition}
	Josh Cowls, Andreas Tsamados, Mariarosaria Taddeo, and Luciano Floridi.
	\newblock A definition, benchmark and database of ai for social good
	initiatives.
	\newblock {\em Nature Machine Intelligence}, 3(2):111--115, 2021.
	
	\bibitem{cui2023drive}
	Can Cui, Yunsheng Ma, Xu~Cao, Wenqian Ye, and Ziran Wang.
	\newblock Drive as you speak: Enabling human-like interaction with large
	language models in autonomous vehicles.
	\newblock {\em arXiv preprint arXiv:2309.10228}, 2023.
	
	\bibitem{daniel2017thinking}
	Kahneman Daniel.
	\newblock {\em Thinking, fast and slow}.
	\newblock 2017.
	
	\bibitem{de2019mesoscopic}
	Felipe de~Souza, Omer Verbas, and Joshua Auld.
	\newblock Mesoscopic traffic flow model for agent-based simulation.
	\newblock {\em Procedia Computer Science}, 151:858--863, 2019.
	
	\bibitem{de2023emergent}
	I~de~Zarz{\`a}, J~de~Curt{\`o}, Gemma Roig, Pietro Manzoni, and Carlos~T
	Calafate.
	\newblock Emergent cooperation and strategy adaptation in multi-agent systems:
	An extended coevolutionary theory with llms.
	\newblock {\em Electronics}, 12(12):2722, 2023.
	
	\bibitem{gemini}
	Google DeepMind.
	\newblock Introducing gemini: our largest and most capable ai model.
	\newblock \url{https://blog.google/technology/ai/google-gemini-ai }, 12 2023.
	\newblock (Accessed on 07/12/2023).
	
	\bibitem{deguchi2011economics}
	Hiroshi Deguchi.
	\newblock {\em Economics as an agent-based complex system: toward agent-based
		social systems sciences}.
	\newblock Springer Science \& Business Media, 2011.
	
	\bibitem{deng2023mind2web}
	Xiang Deng, Yu~Gu, Boyuan Zheng, Shijie Chen, Samuel Stevens, Boshi Wang, Huan
	Sun, and Yu~Su.
	\newblock Mind2web: Towards a generalist agent for the web.
	\newblock {\em arXiv preprint arXiv:2306.06070}, 2023.
	
	\bibitem{deshpande2023toxicity}
	Ameet Deshpande, Vishvak Murahari, Tanmay Rajpurohit, Ashwin Kalyan, and
	Karthik Narasimhan.
	\newblock Toxicity in chatgpt: Analyzing persona-assigned language models.
	\newblock {\em arXiv preprint arXiv:2304.05335}, 2023.
	
	\bibitem{devlin2018bert}
	Jacob Devlin, Ming-Wei Chang, Kenton Lee, and Kristina Toutanova.
	\newblock Bert: Pre-training of deep bidirectional transformers for language
	understanding.
	\newblock {\em arXiv preprint arXiv:1810.04805}, 2018.
	
	\bibitem{dong2022survey}
	Qingxiu Dong, Lei Li, Damai Dai, Ce~Zheng, Zhiyong Wu, Baobao Chang, Xu~Sun,
	Jingjing Xu, and Zhifang Sui.
	\newblock A survey for in-context learning.
	\newblock {\em arXiv preprint arXiv:2301.00234}, 2022.
	
	\bibitem{dubois2023alpacafarm}
	Yann Dubois, Xuechen Li, Rohan Taori, Tianyi Zhang, Ishaan Gulrajani, Jimmy Ba,
	Carlos Guestrin, Percy Liang, and Tatsunori~B Hashimoto.
	\newblock Alpacafarm: A simulation framework for methods that learn from human
	feedback.
	\newblock {\em arXiv preprint arXiv:2305.14387}, 2023.
	
	\bibitem{dwivedi2023benchmarking}
	Vijay~Prakash Dwivedi, Chaitanya~K Joshi, Anh~Tuan Luu, Thomas Laurent, Yoshua
	Bengio, and Xavier Bresson.
	\newblock Benchmarking graph neural networks.
	\newblock {\em Journal of Machine Learning Research}, 24(43):1--48, 2023.
	
	\bibitem{el2012social}
	Abdulrahman~M El-Sayed, Peter Scarborough, Lars Seemann, and Sandro Galea.
	\newblock Social network analysis and agent-based modeling in social
	epidemiology.
	\newblock {\em Epidemiologic Perspectives \& Innovations}, 9(1):1--9, 2012.
	
	\bibitem{elliott2002exploring}
	Euel Elliott and L~Douglas Kiel.
	\newblock Exploring cooperation and competition using agent-based modeling.
	\newblock {\em Proceedings of the National Academy of Sciences},
	99(suppl\_3):7193--7194, 2002.
	
	\bibitem{elsenbroich2014agent}
	Corinna Elsenbroich, Nigel Gilbert, Corinna Elsenbroich, and Nigel Gilbert.
	\newblock Agent-based modelling.
	\newblock {\em Modelling norms}, pages 65--84, 2014.
	
	\bibitem{elyoseph2023chatgpt}
	Zohar Elyoseph, Dorit Hadar-Shoval, Kfir Asraf, and Maya Lvovsky.
	\newblock Chatgpt outperforms humans in emotional awareness evaluations.
	\newblock {\em Frontiers in Psychology}, 14:1199058, 2023.
	
	\bibitem{meta2022human}
	Meta Fundamental AI Research Diplomacy~Team (FAIR)†, Anton Bakhtin, Noam
	Brown, Emily Dinan, Gabriele Farina, Colin Flaherty, Daniel Fried, Andrew
	Goff, Jonathan Gray, Hengyuan Hu, et~al.
	\newblock Human-level play in the game of diplomacy by combining language
	models with strategic reasoning.
	\newblock {\em Science}, 378(6624):1067--1074, 2022.
	
	\bibitem{faria2023artificial}
	Miguel Faria-e Castro and Fernando Leibovici.
	\newblock Artificial intelligence and inflation forecasts.
	\newblock Technical report, 2023.
	
	\bibitem{feng2020learning}
	Jie Feng, Zeyu Yang, Fengli Xu, Haisu Yu, Mudan Wang, and Yong Li.
	\newblock Learning to simulate human mobility.
	\newblock In {\em Proceedings of the 26th ACM SIGKDD international conference
		on knowledge discovery \& data mining}, pages 3426--3433, 2020.
	
	\bibitem{feng2012linking}
	Ling Feng, Baowen Li, Boris Podobnik, Tobias Preis, and H~Eugene Stanley.
	\newblock Linking agent-based models and stochastic models of financial
	markets.
	\newblock {\em Proceedings of the National Academy of Sciences},
	109(22):8388--8393, 2012.
	
	\bibitem{franceschelli2023creativity}
	Giorgio Franceschelli and Mirco Musolesi.
	\newblock On the creativity of large language models.
	\newblock {\em arXiv preprint arXiv:2304.00008}, 2023.
	
	\bibitem{fu2023drive}
	Daocheng Fu, Xin Li, Licheng Wen, Min Dou, Pinlong Cai, Botian Shi, and
	Yu~Qiao.
	\newblock Drive like a human: Rethinking autonomous driving with large language
	models.
	\newblock {\em arXiv preprint arXiv:2307.07162}, 2023.
	
	\bibitem{gao2023s}
	Chen Gao, Xiaochong Lan, Zhihong Lu, Jinzhu Mao, Jinghua Piao, Huandong Wang,
	Depeng Jin, and Yong Li.
	\newblock S3: Social-network simulation system with large language
	model-empowered agents.
	\newblock {\em arXiv preprint arXiv:2307.14984}, 2023.
	
	\bibitem{gatti2011macroeconomics}
	Domenico~Delli Gatti, Saul Desiderio, Edoardo Gaffeo, Pasquale Cirillo, and
	Mauro Gallegati.
	\newblock {\em Macroeconomics from the Bottom-up}, volume~1.
	\newblock Springer Science \& Business Media, 2011.
	
	\bibitem{gaube2013impact}
	Veronika Gaube and Alexander Remesch.
	\newblock Impact of urban planning on household's residential decisions: An
	agent-based simulation model for vienna.
	\newblock {\em Environmental Modelling \& Software}, 45:92--103, 2013.
	
	\bibitem{geerling2023chatgpt}
	Wayne Geerling, G~Dirk Mateer, Jadrian Wooten, and Nikhil Damodaran.
	\newblock Chatgpt has aced the test of understanding in college economics: Now
	what?
	\newblock {\em The American Economist}, page 05694345231169654, 2023.
	
	\bibitem{gilbert2004agent}
	Nigel Gilbert.
	\newblock Agent-based social simulation: dealing with complexity.
	\newblock {\em The Complex Systems Network of Excellence}, 9(25):1--14, 2004.
	
	\bibitem{gilbert2007computational}
	Nigel Gilbert.
	\newblock Computational social science: Agent-based social simulation.
	\newblock In {\em Agent-based modelling and simulation}, pages 115--134.
	Bardwell, 2007.
	
	\bibitem{gilbert2000build}
	Nigel Gilbert and Pietro Terna.
	\newblock How to build and use agent-based models in social science.
	\newblock {\em Mind \& Society}, 1:57--72, 2000.
	
	\bibitem{gilbert2005simulation}
	Nigel Gilbert and Klaus Troitzsch.
	\newblock {\em Simulation for the social scientist}.
	\newblock McGraw-Hill Education (UK), 2005.
	
	\bibitem{guo2023gpt}
	Fulin Guo.
	\newblock Gpt agents in game theory experiments.
	\newblock {\em arXiv preprint arXiv:2305.05516}, 2023.
	
	\bibitem{guo2023suspicion}
	Jiaxian Guo, Bo~Yang, Paul Yoo, Bill~Yuchen Lin, Yusuke Iwasawa, and Yutaka
	Matsuo.
	\newblock Suspicion-agent: Playing imperfect information games with theory of
	mind aware gpt-4.
	\newblock {\em arXiv preprint arXiv:2309.17277}, 2023.
	
	\bibitem{gur2023real}
	Izzeddin Gur, Hiroki Furuta, Austin Huang, Mustafa Safdari, Yutaka Matsuo,
	Douglas Eck, and Aleksandra Faust.
	\newblock A real-world webagent with planning, long context understanding, and
	program synthesis.
	\newblock {\em arXiv preprint arXiv:2307.12856}, 2023.
	
	\bibitem{gurnee2023language}
	Wes Gurnee and Max Tegmark.
	\newblock Language models represent space and time.
	\newblock {\em arXiv preprint arXiv:2310.02207}, 2023.
	
	\bibitem{guyot2006agent}
	Paul Guyot and Shinichi Honiden.
	\newblock Agent-based participatory simulations: Merging multi-agent systems
	and role-playing games.
	\newblock {\em Journal of artificial societies and social simulation}, 9(4),
	2006.
	
	\bibitem{hagendorff2023human}
	Thilo Hagendorff, Sarah Fabi, and Michal Kosinski.
	\newblock Human-like intuitive behavior and reasoning biases emerged in large
	language models but disappeared in chatgpt.
	\newblock {\em Nature Computational Science}, pages 1--6, 2023.
	
	\bibitem{hamalainen2023evaluating}
	Perttu H{\"a}m{\"a}l{\"a}inen, Mikke Tavast, and Anton Kunnari.
	\newblock Evaluating large language models in generating synthetic hci research
	data: a case study.
	\newblock In {\em Proceedings of the 2023 CHI Conference on Human Factors in
		Computing Systems}, pages 1--19, 2023.
	
	\bibitem{hamill2015agent}
	Lynne Hamill and Nigel Gilbert.
	\newblock {\em Agent-based modelling in economics}.
	\newblock John Wiley \& Sons, 2015.
	
	\bibitem{han2023guinea}
	Xu~Han, Zengqing Wu, and Chuan Xiao.
	\newblock " guinea pig trials" utilizing gpt: A novel smart agent-based
	modeling approach for studying firm competition and collusion.
	\newblock {\em arXiv preprint arXiv:2308.10974}, 2023.
	
	\bibitem{hauser2002faculty}
	Marc~D Hauser, Noam Chomsky, and W~Tecumseh Fitch.
	\newblock The faculty of language: what is it, who has it, and how did it
	evolve?
	\newblock {\em science}, 298(5598):1569--1579, 2002.
	
	\bibitem{heckbert2010agent}
	Scott Heckbert, Tim Baynes, and Andrew Reeson.
	\newblock Agent-based modeling in ecological economics.
	\newblock {\em Annals of the New York Academy of Sciences}, 1185(1):39--53,
	2010.
	
	\bibitem{helbing2012social}
	Dirk Helbing.
	\newblock {\em Social self-organization: Agent-based simulations and
		experiments to study emergent social behavior}.
	\newblock Springer, 2012.
	
	\bibitem{helbing1995social}
	Dirk Helbing and Peter Molnar.
	\newblock Social force model for pedestrian dynamics.
	\newblock {\em Physical review E}, 51(5):4282, 1995.
	
	\bibitem{hernandez2016computer}
	Jos{\'e} Hern{\'a}ndez-Orallo, Fernando Mart{\'\i}nez-Plumed, Ute Schmid,
	Michael Siebers, and David~L Dowe.
	\newblock Computer models solving intelligence test problems: Progress and
	implications.
	\newblock {\em Artificial Intelligence}, 230:74--107, 2016.
	
	\bibitem{hong2023metagpt}
	Sirui Hong, Xiawu Zheng, Jonathan Chen, Yuheng Cheng, Ceyao Zhang, Zili Wang,
	Steven Ka~Shing Yau, Zijuan Lin, Liyang Zhou, Chenyu Ran, et~al.
	\newblock Metagpt: Meta programming for multi-agent collaborative framework.
	\newblock {\em arXiv preprint arXiv:2308.00352}, 2023.
	
	\bibitem{horton2023large}
	John~J Horton.
	\newblock Large language models as simulated economic agents: What can we learn
	from homo silicus?
	\newblock Technical report, National Bureau of Economic Research, 2023.
	
	\bibitem{hoshen2017iq}
	Dokhyam Hoshen and Michael Werman.
	\newblock Iq of neural networks.
	\newblock {\em arXiv preprint arXiv:1710.01692}, 2017.
	
	\bibitem{hu2021lora}
	Edward~J Hu, Yelong Shen, Phillip Wallis, Zeyuan Allen-Zhu, Yuanzhi Li, Shean
	Wang, Lu~Wang, and Weizhu Chen.
	\newblock Lora: Low-rank adaptation of large language models.
	\newblock {\em arXiv preprint arXiv:2106.09685}, 2021.
	
	\bibitem{hu2020open}
	Weihua Hu, Matthias Fey, Marinka Zitnik, Yuxiao Dong, Hongyu Ren, Bowen Liu,
	Michele Catasta, and Jure Leskovec.
	\newblock Open graph benchmark: Datasets for machine learning on graphs.
	\newblock {\em Advances in neural information processing systems},
	33:22118--22133, 2020.
	
	\bibitem{hua2023war}
	Wenyue Hua, Lizhou Fan, Lingyao Li, Kai Mei, Jianchao Ji, Yingqiang Ge, Libby
	Hemphill, and Yongfeng Zhang.
	\newblock War and peace (waragent): Large language model-based multi-agent
	simulation of world wars.
	\newblock {\em arXiv preprint arXiv:2311.17227}, 2023.
	
	\bibitem{huang2023benchmarking}
	Qian Huang, Jian Vora, Percy Liang, and Jure Leskovec.
	\newblock Benchmarking large language models as ai research agents.
	\newblock {\em arXiv preprint arXiv:2310.03302}, 2023.
	
	\bibitem{jang2023personalized}
	Joel Jang, Seungone Kim, Bill~Yuchen Lin, Yizhong Wang, Jack Hessel, Luke
	Zettlemoyer, Hannaneh Hajishirzi, Yejin Choi, and Prithviraj Ammanabrolu.
	\newblock Personalized soups: Personalized large language model alignment via
	post-hoc parameter merging.
	\newblock {\em arXiv preprint arXiv:2310.11564}, 2023.
	
	\bibitem{jiang2023health}
	Lavender~Yao Jiang, Xujin~Chris Liu, Nima~Pour Nejatian, Mustafa Nasir-Moin,
	Duo Wang, Anas Abidin, Kevin Eaton, Howard~Antony Riina, Ilya Laufer, Paawan
	Punjabi, et~al.
	\newblock Health system-scale language models are all-purpose prediction
	engines.
	\newblock {\em Nature}, pages 1--6, 2023.
	
	\bibitem{jin2023surrealdriver}
	Ye~Jin, Xiaoxi Shen, Huiling Peng, Xiaoan Liu, Jingli Qin, Jiayang Li, Jintao
	Xie, Peizhong Gao, Guyue Zhou, and Jiangtao Gong.
	\newblock Surrealdriver: Designing generative driver agent simulation framework
	in urban contexts based on large language model.
	\newblock {\em arXiv preprint arXiv:2309.13193}, 2023.
	
	\bibitem{jinxin2023cgmi}
	Shi Jinxin, Zhao Jiabao, Wang Yilei, Wu~Xingjiao, Li~Jiawen, and He~Liang.
	\newblock Cgmi: Configurable general multi-agent interaction framework.
	\newblock {\em arXiv preprint arXiv:2308.12503}, 2023.
	
	\bibitem{kahneman1986fairness}
	Daniel Kahneman, Jack~L Knetsch, and Richard Thaler.
	\newblock Fairness as a constraint on profit seeking: Entitlements in the
	market.
	\newblock {\em The American economic review}, pages 728--741, 1986.
	
	\bibitem{kavak2018big}
	Hamdi Kavak, Jose~J Padilla, Christopher~J Lynch, and Saikou~Y Diallo.
	\newblock Big data, agents, and machine learning: towards a data-driven
	agent-based modeling approach.
	\newblock In {\em Proceedings of the Annual Simulation Symposium}, pages 1--12,
	2018.
	
	\bibitem{kim2021automatic}
	Dongjun Kim, Tae-Sub Yun, Il-Chul Moon, and Jang~Won Bae.
	\newblock Automatic calibration of dynamic and heterogeneous parameters in
	agent-based models.
	\newblock {\em Autonomous Agents and Multi-Agent Systems}, 35(2):46, 2021.
	
	\bibitem{kim2023ai}
	Junsol Kim and Byungkyu Lee.
	\newblock Ai-augmented surveys: Leveraging large language models for opinion
	prediction in nationally representative surveys.
	\newblock {\em arXiv preprint arXiv:2305.09620}, 2023.
	
	\bibitem{kountouriotis2014agent}
	Vassilios Kountouriotis, Stelios~CA Thomopoulos, and Yiannis Papelis.
	\newblock An agent-based crowd behaviour model for real time crowd behaviour
	simulation.
	\newblock {\em Pattern Recognition Letters}, 44:30--38, 2014.
	
	\bibitem{kovavc2023socialai}
	Grgur Kova{\v{c}}, R{\'e}my Portelas, Peter~Ford Dominey, and Pierre-Yves
	Oudeyer.
	\newblock The socialai school: Insights from developmental psychology towards
	artificial socio-cultural agents.
	\newblock {\em arXiv preprint arXiv:2307.07871}, 2023.
	
	\bibitem{kumar2023certifying}
	Aounon Kumar, Chirag Agarwal, Suraj Srinivas, Soheil Feizi, and Hima Lakkaraju.
	\newblock Certifying llm safety against adversarial prompting.
	\newblock {\em arXiv preprint arXiv:2309.02705}, 2023.
	
	\bibitem{lan2023stance}
	Xiaochong Lan, Chen Gao, Depeng Jin, and Yong Li.
	\newblock Stance detection with collaborative role-infused llm-based agents.
	\newblock {\em arXiv preprint arXiv:2310.10467}, 2023.
	
	\bibitem{lengnick2013agent}
	Matthias Lengnick.
	\newblock Agent-based macroeconomics: A baseline model.
	\newblock {\em Journal of Economic Behavior \& Organization}, 86:102--120,
	2013.
	
	\bibitem{leombruni2005economists}
	Roberto Leombruni and Matteo Richiardi.
	\newblock Why are economists sceptical about agent-based simulations?
	\newblock {\em Physica A: Statistical Mechanics and its Applications},
	355(1):103--109, 2005.
	
	\bibitem{li2023seed}
	Bohao Li, Rui Wang, Guangzhi Wang, Yuying Ge, Yixiao Ge, and Ying Shan.
	\newblock Seed-bench: Benchmarking multimodal llms with generative
	comprehension.
	\newblock {\em arXiv preprint arXiv:2307.16125}, 2023.
	
	\bibitem{li2023quantifying}
	Chao Li, Xing Su, Chao Fan, Haoying Han, Cong Xue, and Chunmo Zheng.
	\newblock Quantifying the impact of large language models on collective opinion
	dynamics.
	\newblock {\em arXiv preprint arXiv:2308.03313}, 2023.
	
	\bibitem{li2023modelscope}
	Chenliang Li, Hehong Chen, Ming Yan, Weizhou Shen, Haiyang Xu, Zhikai Wu,
	Zhicheng Zhang, Wenmeng Zhou, Yingda Chen, Chen Cheng, et~al.
	\newblock Modelscope-agent: Building your customizable agent system with
	open-source large language models.
	\newblock {\em arXiv preprint arXiv:2309.00986}, 2023.
	
	\bibitem{li2023camel}
	Guohao Li, Hasan Abed Al~Kader Hammoud, Hani Itani, Dmitrii Khizbullin, and
	Bernard Ghanem.
	\newblock Camel: Communicative agents for" mind" exploration of large scale
	language model society.
	\newblock {\em arXiv preprint arXiv:2303.17760}, 2023.
	
	\bibitem{li2020opinion}
	Ke~Li, Haiming Liang, Gang Kou, and Yucheng Dong.
	\newblock Opinion dynamics model based on the cognitive dissonance: An
	agent-based simulation.
	\newblock {\em Information Fusion}, 56:1--14, 2020.
	
	\bibitem{li2023large}
	Nian Li, Chen Gao, Yong Li, and Qingmin Liao.
	\newblock Large language model-empowered agents for simulating macroeconomic
	activities.
	\newblock {\em arXiv preprint arXiv:2310.10436}, 2023.
	
	\bibitem{li2023you}
	Siyu Li, Jin Yang, and Kui Zhao.
	\newblock Are you in a masquerade? exploring the behavior and impact of large
	language model driven social bots in online social networks.
	\newblock {\em arXiv preprint arXiv:2307.10337}, 2023.
	
	\bibitem{li2022gpt}
	Xingxuan Li, Yutong Li, Linlin Liu, Lidong Bing, and Shafiq Joty.
	\newblock Is gpt-3 a psychopath? evaluating large language models from a
	psychological perspective.
	\newblock {\em arXiv preprint arXiv:2212.10529}, 2022.
	
	\bibitem{liang2023encouraging}
	Tian Liang, Zhiwei He, Wenxiang Jiao, Xing Wang, Yan Wang, Rui Wang, Yujiu
	Yang, Zhaopeng Tu, and Shuming Shi.
	\newblock Encouraging divergent thinking in large language models through
	multi-agent debate.
	\newblock {\em arXiv preprint arXiv:2305.19118}, 2023.
	
	\bibitem{lin2023agentsims}
	Jiaju Lin, Haoran Zhao, Aochi Zhang, Yiting Wu, Huqiuyue Ping, and Qin Chen.
	\newblock Agentsims: An open-source sandbox for large language model
	evaluation.
	\newblock {\em arXiv preprint arXiv:2308.04026}, 2023.
	
	\bibitem{lin1992self}
	Long-Ji Lin.
	\newblock Self-improving reactive agents based on reinforcement learning,
	planning and teaching.
	\newblock {\em Machine learning}, 8:293--321, 1992.
	
	\bibitem{lippe2019using}
	Melvin Lippe, Mike Bithell, Nick Gotts, Davide Natalini, Peter
	Barbrook-Johnson, Carlo Giupponi, Mareen Hallier, Gert~Jan Hofstede,
	Christophe Le~Page, Robin~B Matthews, et~al.
	\newblock Using agent-based modelling to simulate social-ecological systems
	across scales.
	\newblock {\em GeoInformatica}, 23:269--298, 2019.
	
	\bibitem{liu2023training}
	Ruibo Liu, Ruixin Yang, Chenyan Jia, Ge~Zhang, Denny Zhou, Andrew~M Dai, Diyi
	Yang, and Soroush Vosoughi.
	\newblock Training socially aligned language models in simulated human society.
	\newblock {\em arXiv preprint arXiv:2305.16960}, 2023.
	
	\bibitem{liu2023agentbench}
	Xiao Liu, Hao Yu, Hanchen Zhang, Yifan Xu, Xuanyu Lei, Hanyu Lai, Yu~Gu,
	Hangliang Ding, Kaiwen Men, Kejuan Yang, et~al.
	\newblock Agentbench: Evaluating llms as agents.
	\newblock {\em arXiv preprint arXiv:2308.03688}, 2023.
	
	\bibitem{liu2019well}
	Yusen Liu, Fangyuan He, Haodi Zhang, Guozheng Rao, Zhiyong Feng, and Yi~Zhou.
	\newblock How well do machines perform on iq tests: a comparison study on a
	large-scale dataset.
	\newblock In {\em IJCAI}, pages 6110--6116, 2019.
	
	\bibitem{lopez2018microscopic}
	Pablo~Alvarez Lopez, Michael Behrisch, Laura Bieker-Walz, Jakob Erdmann,
	Yun-Pang Fl{\"o}tter{\"o}d, Robert Hilbrich, Leonhard L{\"u}cken, Johannes
	Rummel, Peter Wagner, and Evamarie Wie{\ss}ner.
	\newblock Microscopic traffic simulation using sumo.
	\newblock In {\em 2018 21st international conference on intelligent
		transportation systems (ITSC)}, pages 2575--2582. IEEE, 2018.
	
	\bibitem{lu2023self}
	Jianqiao Lu, Wanjun Zhong, Wenyong Huang, Yufei Wang, Fei Mi, Baojun Wang,
	Weichao Wang, Lifeng Shang, and Qun Liu.
	\newblock Self: Language-driven self-evolution for large language model.
	\newblock {\em arXiv preprint arXiv:2310.00533}, 2023.
	
	\bibitem{luo2008agent}
	Linbo Luo, Suiping Zhou, Wentong Cai, Malcolm Yoke~Hean Low, Feng Tian, Yongwei
	Wang, Xian Xiao, and Dan Chen.
	\newblock Agent-based human behavior modeling for crowd simulation.
	\newblock {\em Computer Animation and Virtual Worlds}, 19(3-4):271--281, 2008.
	
	\bibitem{ma2013agent}
	Yan Ma, SHEN Zhenjiang, and Mitsuhiko Kawakami.
	\newblock Agent-based simulation of residential promoting policy effects on
	downtown revitalization.
	\newblock {\em Journal of Artificial Societies and Social Simulation}, 16(2):2,
	2013.
	
	\bibitem{macal2005tutorial}
	Charles~M Macal and Michael~J North.
	\newblock Tutorial on agent-based modeling and simulation.
	\newblock In {\em Proceedings of the Winter Simulation Conference, 2005.},
	pages 14--pp. IEEE, 2005.
	
	\bibitem{macy2002factors}
	Michael~W Macy and Robert Willer.
	\newblock From factors to actors: Computational sociology and agent-based
	modeling.
	\newblock {\em Annual review of sociology}, 28(1):143--166, 2002.
	
	\bibitem{madey2003agent}
	Gregory Madey, Yongqin Gao, Vincent Freeh, Renee Tynan, and Chris Hoffman.
	\newblock Agent-based modeling and simulation of collaborative social networks.
	\newblock 2003.
	
	\bibitem{maggi2016understanding}
	Elena Maggi and Elena Vallino.
	\newblock Understanding urban mobility and the impact of public policies: The
	role of the agent-based models.
	\newblock {\em Research in Transportation Economics}, 55:50--59, 2016.
	
	\bibitem{mandziuk2019deepiq}
	Jacek Ma{\'n}dziuk and Adam {\.Z}ychowski.
	\newblock Deepiq: A human-inspired ai system for solving iq test problems.
	\newblock In {\em 2019 International Joint Conference on Neural Networks
		(IJCNN)}, pages 1--8. IEEE, 2019.
	
	\bibitem{manvi2023geollm}
	Rohin Manvi, Samar Khanna, Gengchen Mai, Marshall Burke, David Lobell, and
	Stefano Ermon.
	\newblock Geollm: Extracting geospatial knowledge from large language models.
	\newblock {\em arXiv preprint arXiv:2310.06213}, 2023.
	
	\bibitem{mao2023alympics}
	Shaoguang Mao, Yuzhe Cai, Yan Xia, Wenshan Wu, Xun Wang, Fengyi Wang, Tao Ge,
	and Furu Wei.
	\newblock Alympics: Language agents meet game theory.
	\newblock {\em arXiv preprint arXiv:2311.03220}, 2023.
	
	\bibitem{mastio2018distributed}
	Matthieu Mastio, Mahdi Zargayouna, Gerard Scemama, and Omer Rana.
	\newblock Distributed agent-based traffic simulations.
	\newblock {\em IEEE Intelligent Transportation Systems Magazine},
	10(1):145--156, 2018.
	
	\bibitem{mclane2011role}
	Adam~J McLane, Christina Semeniuk, Gregory~J McDermid, and Danielle~J Marceau.
	\newblock The role of agent-based models in wildlife ecology and management.
	\newblock {\em Ecological modelling}, 222(8):1544--1556, 2011.
	
	\bibitem{melis2017state}
	G{\'a}bor Melis, Chris Dyer, and Phil Blunsom.
	\newblock On the state of the art of evaluation in neural language models.
	\newblock {\em arXiv preprint arXiv:1707.05589}, 2017.
	
	\bibitem{mueller2016economic}
	Matthias Mueller and Andreas Pyka.
	\newblock Economic behaviour and agent-based modelling.
	\newblock In {\em Routledge handbook of behavioral economics}, pages 405--415.
	Routledge, 2016.
	
	\bibitem{mukobi2023welfare}
	Gabriel Mukobi, Hannah Erlebach, Niklas Lauffer, Lewis Hammond, Alan Chan, and
	Jesse Clifton.
	\newblock Welfare diplomacy: Benchmarking language model cooperation.
	\newblock {\em arXiv preprint arXiv:2310.08901}, 2023.
	
	\bibitem{nascimento2023self}
	Nathalia Nascimento, Paulo Alencar, and Donald Cowan.
	\newblock Self-adaptive large language model (llm)-based multiagent systems.
	\newblock {\em arXiv preprint arXiv:2307.06187}, 2023.
	
	\bibitem{national2018open}
	Engineering National Academies~of Sciences, Medicine, et~al.
	\newblock Open science by design: Realizing a vision for 21st century research.
	\newblock 2018.
	
	\bibitem{chatgpt}
	OpenAI.
	\newblock Introducing chatgpt.
	\newblock \url{https://openai.com/blog/chatgpt}, 2022.
	\newblock (Accessed on 01/12/2023).
	
	\bibitem{osoba2020policy}
	Osonde~A Osoba, Raffaele Vardavas, Justin Grana, Rushil Zutshi, and Amber
	Jaycocks.
	\newblock Policy-focused agent-based modeling using rl behavioral models.
	\newblock {\em arXiv preprint arXiv:2006.05048}, 2020.
	
	\bibitem{park2023choicemates}
	Jeongeon Park, Bryan Min, Xiaojuan Ma, and Juho Kim.
	\newblock Choicemates: Supporting unfamiliar online decision-making with
	multi-agent conversational interactions.
	\newblock {\em arXiv preprint arXiv:2310.01331}, 2023.
	
	\bibitem{park2023generative}
	Joon~Sung Park, Joseph O'Brien, Carrie~Jun Cai, Meredith~Ringel Morris, Percy
	Liang, and Michael~S Bernstein.
	\newblock Generative agents: Interactive simulacra of human behavior.
	\newblock In {\em Proceedings of the 36th Annual ACM Symposium on User
		Interface Software and Technology}, pages 1--22, 2023.
	
	\bibitem{park2022social}
	Joon~Sung Park, Lindsay Popowski, Carrie Cai, Meredith~Ringel Morris, Percy
	Liang, and Michael~S Bernstein.
	\newblock Social simulacra: Creating populated prototypes for social computing
	systems.
	\newblock In {\em Proceedings of the 35th Annual ACM Symposium on User
		Interface Software and Technology}, pages 1--18, 2022.
	
	\bibitem{park2019multi}
	Young~Joon Park, Yoon~Sang Cho, and Seoung~Bum Kim.
	\newblock Multi-agent reinforcement learning with approximate model learning
	for competitive games.
	\newblock {\em PloS one}, 14(9):e0222215, 2019.
	
	\bibitem{parv2019agent}
	Luminita Parv, Bogdan Deaky, Marius~Daniel Nasulea, and Gheorghe Oancea.
	\newblock Agent-based simulation of value flow in an industrial production
	process.
	\newblock {\em Processes}, 7(2):82, 2019.
	
	\bibitem{pereira2004agent}
	Ant{\'o}nio Pereira, Pedro Duarte, and Lu{\'\i}s~Paulo Reis.
	\newblock Agent-based simulation of ecological models.
	\newblock In {\em Agent-Based Simulation}, 2004.
	
	\bibitem{perez2009agent}
	Liliana Perez and Suzana Dragicevic.
	\newblock An agent-based approach for modeling dynamics of contagious disease
	spread.
	\newblock {\em International journal of health geographics}, 8(1):1--17, 2009.
	
	\bibitem{pertoldi2004impact}
	Cino Pertoldi and Chris Topping.
	\newblock Impact assessment predicted by means of genetic agent-based modeling.
	\newblock {\em Critical reviews in toxicology}, 34(6):487--498, 2004.
	
	\bibitem{phelps2023investigating}
	Steve Phelps and Yvan~I Russell.
	\newblock Investigating emergent goal-like behaviour in large language models
	using experimental economics.
	\newblock {\em arXiv preprint arXiv:2305.07970}, 2023.
	
	\bibitem{platas2023survey}
	Alejandro Platas-L{\'o}pez, Alejandro Guerra-Hern{\'a}ndez, Marcela
	Quiroz-Castellanos, and Nicandro Cruz-Ramirez.
	\newblock A survey on agent-based modelling assisted by machine learning.
	\newblock {\em Expert Systems}, page e13325, 2023.
	
	\bibitem{plosser1979potential}
	Charles~I Plosser and G~William Schwert.
	\newblock Potential gnp: Its measurement and significance: A dissenting
	opinion.
	\newblock In {\em Carnegie-Rochester Conference Series on Public Policy},
	volume~10, pages 179--186. Elsevier, 1979.
	
	\bibitem{puranam2015modelling}
	Phanish Puranam, Nils Stieglitz, Magda Osman, and Madan~M Pillutla.
	\newblock Modelling bounded rationality in organizations: Progress and
	prospects.
	\newblock {\em Academy of Management Annals}, 9(1):337--392, 2015.
	
	\bibitem{qian2023communicative}
	Chen Qian, Xin Cong, Cheng Yang, Weize Chen, Yusheng Su, Juyuan Xu, Zhiyuan
	Liu, and Maosong Sun.
	\newblock Communicative agents for software development.
	\newblock {\em arXiv preprint arXiv:2307.07924}, 2023.
	
	\bibitem{qin2023tool}
	Yujia Qin, Shengding Hu, Yankai Lin, Weize Chen, Ning Ding, Ganqu Cui, Zheni
	Zeng, Yufei Huang, Chaojun Xiao, Chi Han, et~al.
	\newblock Tool learning with foundation models.
	\newblock {\em arXiv preprint arXiv:2304.08354}, 2023.
	
	\bibitem{radford2019language}
	Alec Radford, Jeffrey Wu, Rewon Child, David Luan, Dario Amodei, Ilya
	Sutskever, et~al.
	\newblock Language models are unsupervised multitask learners.
	\newblock {\em OpenAI blog}, 1(8):9, 2019.
	
	\bibitem{rolon2012agent}
	Milagros Rol{\'o}n and Ernesto Mart{\'\i}nez.
	\newblock Agent-based modeling and simulation of an autonomic manufacturing
	execution system.
	\newblock {\em Computers in industry}, 63(1):53--78, 2012.
	
	\bibitem{rouchier2017agent}
	Juliette Rouchier.
	\newblock Agent-based simulation as a useful tool for the study of markets.
	\newblock {\em Simulating Social Complexity: A Handbook}, pages 671--704, 2017.
	
	\bibitem{russakovsky2015imagenet}
	Olga Russakovsky, Jia Deng, Hao Su, Jonathan Krause, Sanjeev Satheesh, Sean Ma,
	Zhiheng Huang, Andrej Karpathy, Aditya Khosla, Michael Bernstein, et~al.
	\newblock Imagenet large scale visual recognition challenge.
	\newblock {\em International journal of computer vision}, 115:211--252, 2015.
	
	\bibitem{samanidou2007agent}
	Egle Samanidou, Elmar Zschischang, Dietrich Stauffer, and Thomas Lux.
	\newblock Agent-based models of financial markets.
	\newblock {\em Reports on Progress in Physics}, 70(3):409, 2007.
	
	\bibitem{samuelson1988status}
	William Samuelson and Richard Zeckhauser.
	\newblock Status quo bias in decision making.
	\newblock {\em Journal of risk and uncertainty}, 1:7--59, 1988.
	
	\bibitem{schick2023toolformer}
	Timo Schick, Jane Dwivedi-Yu, Roberto Dess{\`\i}, Roberta Raileanu, Maria
	Lomeli, Luke Zettlemoyer, Nicola Cancedda, and Thomas Scialom.
	\newblock Toolformer: Language models can teach themselves to use tools.
	\newblock {\em arXiv preprint arXiv:2302.04761}, 2023.
	
	\bibitem{schieritz2003emergent}
	Nadine Schieritz and Andreas Grobler.
	\newblock Emergent structures in supply chains-a study integrating agent-based
	and system dynamics modeling.
	\newblock In {\em 36th Annual Hawaii International Conference on System
		Sciences, 2003. Proceedings of the}, pages 9--pp. IEEE, 2003.
	
	\bibitem{schwitzgebel2023creating}
	Eric Schwitzgebel, David Schwitzgebel, and Anna Strasser.
	\newblock Creating a large language model of a philosopher.
	\newblock {\em arXiv preprint arXiv:2302.01339}, 2023.
	
	\bibitem{sert2020segregation}
	Egemen Sert, Yaneer Bar-Yam, and Alfredo~J Morales.
	\newblock Segregation dynamics with reinforcement learning and agent based
	modeling.
	\newblock {\em Scientific reports}, 10(1):11771, 2020.
	
	\bibitem{shah2023lm}
	Dhruv Shah, B{\l}a{\.z}ej Osi{\'n}ski, Sergey Levine, et~al.
	\newblock Lm-nav: Robotic navigation with large pre-trained models of language,
	vision, and action.
	\newblock In {\em Conference on Robot Learning}, pages 492--504. PMLR, 2023.
	
	\bibitem{shah2023gnm}
	Dhruv Shah, Ajay Sridhar, Arjun Bhorkar, Noriaki Hirose, and Sergey Levine.
	\newblock Gnm: A general navigation model to drive any robot.
	\newblock In {\em 2023 IEEE International Conference on Robotics and Automation
		(ICRA)}, pages 7226--7233. IEEE, 2023.
	
	\bibitem{shaikh2023rehearsal}
	Omar Shaikh, Valentino Chai, Michele~J Gelfand, Diyi Yang, and Michael~S
	Bernstein.
	\newblock Rehearsal: Simulating conflict to teach conflict resolution.
	\newblock {\em arXiv preprint arXiv:2309.12309}, 2023.
	
	\bibitem{shanahan2023role}
	Murray Shanahan, Kyle McDonell, and Laria Reynolds.
	\newblock Role play with large language models.
	\newblock {\em Nature}, pages 1--6, 2023.
	
	\bibitem{shen2021towards}
	Zheyan Shen, Jiashuo Liu, Yue He, Xingxuan Zhang, Renzhe Xu, Han Yu, and Peng
	Cui.
	\newblock Towards out-of-distribution generalization: A survey.
	\newblock {\em arXiv preprint arXiv:2108.13624}, 2021.
	
	\bibitem{sheng2023high}
	Ying Sheng, Lianmin Zheng, Binhang Yuan, Zhuohan Li, Max Ryabinin, Daniel~Y Fu,
	Zhiqiang Xie, Beidi Chen, Clark Barrett, Joseph~E Gonzalez, et~al.
	\newblock High-throughput generative inference of large language models with a
	single gpu.
	\newblock {\em arXiv preprint arXiv:2303.06865}, 2023.
	
	\bibitem{shinn2023reflexion}
	Noah Shinn, Federico Cassano, Ashwin Gopinath, Karthik~R Narasimhan, and Shunyu
	Yao.
	\newblock Reflexion: Language agents with verbal reinforcement learning.
	\newblock In {\em Thirty-seventh Conference on Neural Information Processing
		Systems}, 2023.
	
	\bibitem{shridhar2020alfworld}
	Mohit Shridhar, Xingdi Yuan, Marc-Alexandre C{\^o}t{\'e}, Yonatan Bisk, Adam
	Trischler, and Matthew Hausknecht.
	\newblock Alfworld: Aligning text and embodied environments for interactive
	learning.
	\newblock {\em arXiv preprint arXiv:2010.03768}, 2020.
	
	\bibitem{silva2020covid}
	Petr{\^o}nio~CL Silva, Paulo~VC Batista, H{\'e}lder~S Lima, Marcos~A Alves,
	Frederico~G Guimar{\~a}es, and Rodrigo~CP Silva.
	\newblock Covid-abs: An agent-based model of covid-19 epidemic to simulate
	health and economic effects of social distancing interventions.
	\newblock {\em Chaos, Solitons \& Fractals}, 139:110088, 2020.
	
	\bibitem{silverman2015systems}
	Barry~G Silverman, Nancy Hanrahan, Gnana Bharathy, Kim Gordon, and Dan Johnson.
	\newblock A systems approach to healthcare: agent-based modeling, community
	mental health, and population well-being.
	\newblock {\em Artificial intelligence in medicine}, 63(2):61--71, 2015.
	
	\bibitem{simon1997models}
	Herbert~Alexander Simon.
	\newblock {\em Models of bounded rationality: Empirically grounded economic
		reason}, volume~3.
	\newblock MIT press, 1997.
	
	\bibitem{singh2023mind}
	Manmeet Singh, Vaisakh SB, Neetiraj Malviya, et~al.
	\newblock Mind meets machine: Unravelling gpt-4's cognitive psychology.
	\newblock {\em arXiv preprint arXiv:2303.11436}, 2023.
	
	\bibitem{singhal2023large}
	Karan Singhal, Shekoofeh Azizi, Tao Tu, S~Sara Mahdavi, Jason Wei, Hyung~Won
	Chung, Nathan Scales, Ajay Tanwani, Heather Cole-Lewis, Stephen Pfohl, et~al.
	\newblock Large language models encode clinical knowledge.
	\newblock {\em Nature}, 620(7972):172--180, 2023.
	
	\bibitem{song2023llm}
	Chan~Hee Song, Jiaman Wu, Clayton Washington, Brian~M Sadler, Wei-Lun Chao, and
	Yu~Su.
	\newblock Llm-planner: Few-shot grounded planning for embodied agents with
	large language models.
	\newblock In {\em Proceedings of the IEEE/CVF International Conference on
		Computer Vision}, pages 2998--3009, 2023.
	
	\bibitem{sun2023adaplanner}
	Haotian Sun, Yuchen Zhuang, Lingkai Kong, Bo~Dai, and Chao Zhang.
	\newblock Adaplanner: Adaptive planning from feedback with language models.
	\newblock {\em arXiv preprint arXiv:2305.16653}, 2023.
	
	\bibitem{surowiecki2005wisdom}
	James Surowiecki.
	\newblock {\em The wisdom of crowds}.
	\newblock Anchor, 2005.
	
	\bibitem{suzuki2023evolutionary}
	Reiji Suzuki and Takaya Arita.
	\newblock An evolutionary model of personality traits related to cooperative
	behavior using a large language model.
	\newblock {\em arXiv preprint arXiv:2310.05976}, 2023.
	
	\bibitem{alpaca}
	Rohan Taori, Ishaan Gulrajani, Tianyi Zhang, Yann Dubois, Xuechen Li, Carlos
	Guestrin, Percy Liang, and Tatsunori~B. Hashimoto.
	\newblock Stanford alpaca: An instruction-following llama model.
	\newblock \url{https://github.com/tatsu-lab/stanford_alpaca}, 2023.
	
	\bibitem{autogpt}
	AutoGPT Team.
	\newblock Autogpt: the heart of the open-source agent ecosystem.
	\newblock \url{https://github.com/Significant-Gravitas/AutoGPT}, 2022.
	\newblock (Accessed on 01/10/2023).
	
	\bibitem{xagent2023}
	XAgent Team.
	\newblock Xagent: An autonomous agent for complex task solving, 2023.
	
	\bibitem{terna1998simulation}
	Pietro Terna et~al.
	\newblock Simulation tools for social scientists: Building agent based models
	with swarm.
	\newblock {\em Journal of artificial societies and social simulation},
	1(2):1--12, 1998.
	
	\bibitem{thirunavukarasu2023large}
	Arun~James Thirunavukarasu, Darren Shu~Jeng Ting, Kabilan Elangovan, Laura
	Gutierrez, Ting~Fang Tan, and Daniel Shu~Wei Ting.
	\newblock Large language models in medicine.
	\newblock {\em Nature medicine}, 29(8):1930--1940, 2023.
	
	\bibitem{tian2023evil}
	Yu~Tian, Xiao Yang, Jingyuan Zhang, Yinpeng Dong, and Hang Su.
	\newblock Evil geniuses: Delving into the safety of llm-based agents.
	\newblock {\em arXiv preprint arXiv:2311.11855}, 2023.
	
	\bibitem{tomasello2010origins}
	Michael Tomasello.
	\newblock {\em Origins of human communication}.
	\newblock MIT press, 2010.
	
	\bibitem{touvron2023llama}
	Hugo Touvron, Thibaut Lavril, Gautier Izacard, Xavier Martinet, Marie-Anne
	Lachaux, Timoth{\'e}e Lacroix, Baptiste Rozi{\`e}re, Naman Goyal, Eric
	Hambro, Faisal Azhar, et~al.
	\newblock Llama: Open and efficient foundation language models.
	\newblock {\em arXiv preprint arXiv:2302.13971}, 2023.
	
	\bibitem{valmeekam2022large}
	Karthik Valmeekam, Alberto Olmo, Sarath Sreedharan, and Subbarao Kambhampati.
	\newblock Large language models still can't plan (a benchmark for llms on
	planning and reasoning about change).
	\newblock {\em arXiv preprint arXiv:2206.10498}, 2022.
	
	\bibitem{van2008agent}
	Clemens Van~Dinther.
	\newblock Agent-based simulation for research in economics.
	\newblock In {\em Handbook on information technology in finance}, pages
	421--442. Springer, 2008.
	
	\bibitem{vijayaraghavan2023minimum}
	Avish Vijayaraghavan and Cosmin Badea.
	\newblock Minimum levels of interpretability for artificial moral agents.
	\newblock {\em arXiv preprint arXiv:2307.00660}, 2023.
	
	\bibitem{wall2016agent}
	Friederike Wall.
	\newblock Agent-based modeling in managerial science: an illustrative survey
	and study.
	\newblock {\em Review of Managerial Science}, 10(1):135--193, 2016.
	
	\bibitem{wang2018glue}
	Alex Wang, Amanpreet Singh, Julian Michael, Felix Hill, Omer Levy, and Samuel~R
	Bowman.
	\newblock Glue: A multi-task benchmark and analysis platform for natural
	language understanding.
	\newblock {\em arXiv preprint arXiv:1804.07461}, 2018.
	
	\bibitem{wang2023voyager}
	Guanzhi Wang, Yuqi Xie, Yunfan Jiang, Ajay Mandlekar, Chaowei Xiao, Yuke Zhu,
	Linxi Fan, and Anima Anandkumar.
	\newblock Voyager: An open-ended embodied agent with large language models.
	\newblock {\em arXiv preprint arXiv:2305.16291}, 2023.
	
	\bibitem{wang2023robustness}
	Jindong Wang, Xixu Hu, Wenxin Hou, Hao Chen, Runkai Zheng, Yidong Wang, Linyi
	Yang, Haojun Huang, Wei Ye, Xiubo Geng, et~al.
	\newblock On the robustness of chatgpt: An adversarial and out-of-distribution
	perspective.
	\newblock {\em arXiv preprint arXiv:2302.12095}, 2023.
	
	\bibitem{wang2018agent}
	L~Wang, Kwangwon Ahn, C~Kim, and C~Ha.
	\newblock Agent-based models in financial market studies.
	\newblock In {\em Journal of Physics: Conference Series}, volume 1039, page
	012022. IOP Publishing, 2018.
	
	\bibitem{wang2023survey}
	Lei Wang, Chen Ma, Xueyang Feng, Zeyu Zhang, Hao Yang, Jingsen Zhang, Zhiyuan
	Chen, Jiakai Tang, Xu~Chen, Yankai Lin, et~al.
	\newblock A survey on large language model based autonomous agents.
	\newblock {\em arXiv preprint arXiv:2308.11432}, 2023.
	
	\bibitem{wang2023recagent}
	Lei Wang, Jingsen Zhang, Xu~Chen, Yankai Lin, Ruihua Song, Wayne~Xin Zhao, and
	Ji-Rong Wen.
	\newblock Recagent: A novel simulation paradigm for recommender systems.
	\newblock {\em arXiv preprint arXiv:2306.02552}, 2023.
	
	\bibitem{wang2023unleashing}
	Zhenhailong Wang, Shaoguang Mao, Wenshan Wu, Tao Ge, Furu Wei, and Heng Ji.
	\newblock Unleashing cognitive synergy in large language models: A task-solving
	agent through multi-persona self-collaboration.
	\newblock {\em arXiv preprint arXiv:2307.05300}, 2023.
	
	\bibitem{wang2023humanoid}
	Zhilin Wang, Yu~Ying Chiu, and Yu~Cheung Chiu.
	\newblock Humanoid agents: Platform for simulating human-like generative
	agents.
	\newblock {\em arXiv preprint arXiv:2310.05418}, 2023.
	
	\bibitem{weber2004success}
	Steven Weber.
	\newblock {\em The success of open source}.
	\newblock Harvard University Press, 2004.
	
	\bibitem{wei2022chain}
	Jason Wei, Xuezhi Wang, Dale Schuurmans, Maarten Bosma, Fei Xia, Ed~Chi, Quoc~V
	Le, Denny Zhou, et~al.
	\newblock Chain-of-thought prompting elicits reasoning in large language
	models.
	\newblock {\em Advances in Neural Information Processing Systems},
	35:24824--24837, 2022.
	
	\bibitem{widener2013agent}
	Michael~J Widener, Sara~S Metcalf, and Yaneer Bar-Yam.
	\newblock Agent-based modeling of policies to improve urban food access for
	low-income populations.
	\newblock {\em Applied Geography}, 40:1--10, 2013.
	
	\bibitem{williams2023epidemic}
	Ross Williams, Niyousha Hosseinichimeh, Aritra Majumdar, and Navid
	Ghaffarzadegan.
	\newblock Epidemic modeling with generative agents.
	\newblock {\em arXiv preprint arXiv:2307.04986}, 2023.
	
	\bibitem{wolfram1984cellular}
	Stephen Wolfram.
	\newblock Cellular automata as models of complexity.
	\newblock {\em Nature}, 311(5985):419--424, 1984.
	
	\bibitem{wooldridge1995intelligent}
	Michael Wooldridge and Nicholas~R Jennings.
	\newblock Intelligent agents: Theory and practice.
	\newblock {\em The knowledge engineering review}, 10(2):115--152, 1995.
	
	\bibitem{wu2023plan}
	Yue Wu, So~Yeon Min, Yonatan Bisk, Ruslan Salakhutdinov, Amos Azaria, Yuanzhi
	Li, Tom Mitchell, and Shrimai Prabhumoye.
	\newblock Plan, eliminate, and track--language models are good teachers for
	embodied agents.
	\newblock {\em arXiv preprint arXiv:2305.02412}, 2023.
	
	\bibitem{xi2023rise}
	Zhiheng Xi, Wenxiang Chen, Xin Guo, Wei He, Yiwen Ding, Boyang Hong, Ming
	Zhang, Junzhe Wang, Senjie Jin, Enyu Zhou, et~al.
	\newblock The rise and potential of large language model based agents: A
	survey.
	\newblock {\em arXiv preprint arXiv:2309.07864}, 2023.
	
	\bibitem{xie2023wall}
	Qianqian Xie, Weiguang Han, Yanzhao Lai, Min Peng, and Jimin Huang.
	\newblock The wall street neophyte: A zero-shot analysis of chatgpt over
	multimodal stock movement prediction challenges.
	\newblock {\em arXiv preprint arXiv:2304.05351}, 2023.
	
	\bibitem{ugi}
	Fengli Xu, Jun Zhang, Chen Gao, Jie Feng, and Yong Li.
	\newblock Urban generative intelligence (ugi): A foundational platform for
	embodied agent and future city.
	\newblock {\em arXiv preprint}, 2023.
	
	\bibitem{xu2023exploring}
	Yuzhuang Xu, Shuo Wang, Peng Li, Fuwen Luo, Xiaolong Wang, Weidong Liu, and
	Yang Liu.
	\newblock Exploring large language models for communication games: An empirical
	study on werewolf.
	\newblock {\em arXiv preprint arXiv:2309.04658}, 2023.
	
	\bibitem{yao2023instructions}
	Jing Yao, Xiaoyuan Yi, Xiting Wang, Jindong Wang, and Xing Xie.
	\newblock From instructions to intrinsic human values--a survey of alignment
	goals for big models.
	\newblock {\em arXiv preprint arXiv:2308.12014}, 2023.
	
	\bibitem{yao2023tree}
	Shunyu Yao, Dian Yu, Jeffrey Zhao, Izhak Shafran, Thomas~L Griffiths, Yuan Cao,
	and Karthik Narasimhan.
	\newblock Tree of thoughts: Deliberate problem solving with large language
	models.
	\newblock {\em arXiv preprint arXiv:2305.10601}, 2023.
	
	\bibitem{yi2023unpacking}
	Xiaoyuan Yi, Jing Yao, Xiting Wang, and Xing Xie.
	\newblock Unpacking the ethical value alignment in big models.
	\newblock {\em arXiv preprint arXiv:2310.17551}, 2023.
	
	\bibitem{babyagi}
	Yoheinakajima.
	\newblock Babyagi.
	\newblock \url{https://github.com/yoheinakajima/babyagi}, 2023.
	\newblock (Accessed on 01/10/2023).
	
	\bibitem{zeng2023glm}
	Aohan Zeng, Xiao Liu, Zhengxiao Du, Zihan Wang, Hanyu Lai, Ming Ding, Zhuoyi
	Yang, Yifan Xu, Wendi Zheng, Xiao Xia, et~al.
	\newblock Glm-130b: An open bilingual pre-trained model.
	\newblock In {\em The Eleventh International Conference on Learning
		Representations}, 2023.
	
	\bibitem{zhang2023generative}
	An~Zhang, Leheng Sheng, Yuxin Chen, Hao Li, Yang Deng, Xiang Wang, and Tat-Seng
	Chua.
	\newblock On generative agents in recommendation.
	\newblock {\em arXiv preprint arXiv:2310.10108}, 2023.
	
	\bibitem{zhang2020overview}
	Bo~Zhang and Donald~L DeAngelis.
	\newblock An overview of agent-based models in plant biology and ecology.
	\newblock {\em Annals of Botany}, 126(4):539--557, 2020.
	
	\bibitem{zhang2023building}
	Hongxin Zhang, Weihua Du, Jiaming Shan, Qinhong Zhou, Yilun Du, Joshua~B
	Tenenbaum, Tianmin Shu, and Chuang Gan.
	\newblock Building cooperative embodied agents modularly with large language
	models.
	\newblock {\em arXiv preprint arXiv:2307.02485}, 2023.
	
	\bibitem{zhang2023exploring}
	Jintian Zhang, Xin Xu, and Shumin Deng.
	\newblock Exploring collaboration mechanisms for llm agents: A social
	psychology view.
	\newblock {\em arXiv preprint arXiv:2310.02124}, 2023.
	
	\bibitem{zhang2023benchmarking}
	Tianyi Zhang, Faisal Ladhak, Esin Durmus, Percy Liang, Kathleen McKeown, and
	Tatsunori~B Hashimoto.
	\newblock Benchmarking large language models for news summarization.
	\newblock {\em arXiv preprint arXiv:2301.13848}, 2023.
	
	\bibitem{zhao2023explainability}
	Haiyan Zhao, Hanjie Chen, Fan Yang, Ninghao Liu, Huiqi Deng, Hengyi Cai,
	Shuaiqiang Wang, Dawei Yin, and Mengnan Du.
	\newblock Explainability for large language models: A survey.
	\newblock {\em arXiv preprint arXiv:2309.01029}, 2023.
	
	\bibitem{zhao2023competeai}
	Qinlin Zhao, Jindong Wang, Yixuan Zhang, Yiqiao Jin, Kaijie Zhu, Hao Chen, and
	Xing Xie.
	\newblock Competeai: Understanding the competition behaviors in large language
	model-based agents.
	\newblock {\em arXiv preprint arXiv:2310.17512}, 2023.
	
	\bibitem{zhao2023survey}
	Wayne~Xin Zhao, Kun Zhou, Junyi Li, Tianyi Tang, Xiaolei Wang, Yupeng Hou,
	Yingqian Min, Beichen Zhang, Junjie Zhang, Zican Dong, et~al.
	\newblock A survey of large language models.
	\newblock {\em arXiv preprint arXiv:2303.18223}, 2023.
	
	\bibitem{zheng2022ai}
	Stephan Zheng, Alexander Trott, Sunil Srinivasa, David~C Parkes, and Richard
	Socher.
	\newblock The ai economist: Taxation policy design via two-level deep
	multiagent reinforcement learning.
	\newblock {\em Science advances}, 8(18):eabk2607, 2022.
	
	\bibitem{zhou2023webarena}
	Shuyan Zhou, Frank~F Xu, Hao Zhu, Xuhui Zhou, Robert Lo, Abishek Sridhar,
	Xianyi Cheng, Yonatan Bisk, Daniel Fried, Uri Alon, et~al.
	\newblock Webarena: A realistic web environment for building autonomous agents.
	\newblock {\em arXiv preprint arXiv:2307.13854}, 2023.
	
	\bibitem{zhou2023sotopia}
	Xuhui Zhou, Hao Zhu, Leena Mathur, Ruohong Zhang, Haofei Yu, Zhengyang Qi,
	Louis-Philippe Morency, Yonatan Bisk, Daniel Fried, Graham Neubig, et~al.
	\newblock Sotopia: Interactive evaluation for social intelligence in language
	agents.
	\newblock {\em arXiv preprint arXiv:2310.11667}, 2023.
	
	\bibitem{zhu2023minigpt}
	Deyao Zhu, Jun Chen, Xiaoqian Shen, Xiang Li, and Mohamed Elhoseiny.
	\newblock Minigpt-4: Enhancing vision-language understanding with advanced
	large language models.
	\newblock {\em arXiv preprint arXiv:2304.10592}, 2023.
	
	\bibitem{zhu2023promptbench}
	Kaijie Zhu, Jindong Wang, Jiaheng Zhou, Zichen Wang, Hao Chen, Yidong Wang,
	Linyi Yang, Wei Ye, Neil~Zhenqiang Gong, Yue Zhang, et~al.
	\newblock Promptbench: Towards evaluating the robustness of large language
	models on adversarial prompts.
	\newblock {\em arXiv preprint arXiv:2306.04528}, 2023.
	
	\bibitem{zhu2023ghost}
	Xizhou Zhu, Yuntao Chen, Hao Tian, Chenxin Tao, Weijie Su, Chenyu Yang, Gao
	Huang, Bin Li, Lewei Lu, Xiaogang Wang, et~al.
	\newblock Ghost in the minecraft: Generally capable agents for open-world
	enviroments via large language models with text-based knowledge and memory.
	\newblock {\em arXiv preprint arXiv:2305.17144}, 2023.
	
	\bibitem{zhu2023survey}
	Xunyu Zhu, Jian Li, Yong Liu, Can Ma, and Weiping Wang.
	\newblock A survey on model compression for large language models.
	\newblock {\em arXiv preprint arXiv:2308.07633}, 2023.
	
	\bibitem{zhuge2023mindstorms}
	Mingchen Zhuge, Haozhe Liu, Francesco Faccio, Dylan~R Ashley, R{\'o}bert
	Csord{\'a}s, Anand Gopalakrishnan, Abdullah Hamdi, Hasan Abed Al~Kader
	Hammoud, Vincent Herrmann, Kazuki Irie, et~al.
	\newblock Mindstorms in natural language-based societies of mind.
	\newblock {\em arXiv preprint arXiv:2305.17066}, 2023.
	
	\bibitem{zhuo2023exploring}
	Terry~Yue Zhuo, Yujin Huang, Chunyang Chen, and Zhenchang Xing.
	\newblock Exploring ai ethics of chatgpt: A diagnostic analysis.
	\newblock {\em arXiv preprint arXiv:2301.12867}, 2023.
	
	\bibitem{zou2023wireless}
	Hang Zou, Qiyang Zhao, Lina Bariah, Mehdi Bennis, and Merouane Debbah.
	\newblock Wireless multi-agent generative ai: From connected intelligence to
	collective intelligence.
	\newblock {\em arXiv preprint arXiv:2307.02757}, 2023.
	
\end{thebibliography}

\end{document}